\documentclass{article}
\usepackage{multicol}

\usepackage{arxiv}

\usepackage[utf8]{inputenc} % allow utf-8 input
\usepackage[T1]{fontenc}    % use 8-bit T1 fonts
\usepackage{hyperref}       % hyperlinks
\usepackage{url}            % simple URL typesetting
\usepackage{booktabs}       % professional-quality tables
\usepackage{nicefrac}       % compact symbols for 1/2, etc.
\usepackage{microtype}      % microtypography
\usepackage{lipsum}		% Can be removed after putting your text content
\usepackage{graphicx}
\usepackage{doi}
\usepackage{caption}
\usepackage{graphicx}
\usepackage{float} 
\usepackage{color}
\usepackage{subcaption}
\usepackage[ruled,linesnumbered]{algorithm2e}
\usepackage{textcomp}
\usepackage{amsmath,amssymb,amsfonts}
\usepackage{soul}
\soulregister{\cite}7
\soulregister{\ref}7
\renewcommand{\hl}[1]{#1}

\title{Scheduled Curiosity-Deep Dyna-Q: Efficient Exploration for Dialog Policy Learning}

\author{
	Xuecheng Niu, Akinori Ito, Takashi Nose \\
	Graduate School of Engineering \\
	Tohoku University \\
	Sendai, Japan\\
	\texttt{\{niu.xuecheng.p8@dc, aito.spcom@, nose@\}tohoku.ac.jp} \\
}

\renewcommand{\shorttitle}{\textit{arXiv} Template}

\begin{document}
	
	\maketitle
	
	\begin{abstract}
		Training task-oriented dialog agents based on reinforcement learning is time-consuming and requires a large number of interactions with real users. How to grasp dialog policy within limited dialog experiences remains an obstacle that makes the agent training process less efficient. 
		In addition, most previous frameworks start training by randomly choosing training samples, which differs from the human learning method and hurts the efficiency and stability of training. 
		Therefore, we propose Scheduled Curiosity-Deep Dyna-Q (SC-DDQ), a curiosity-driven curriculum learning framework based on a state-of-the-art model-based reinforcement learning dialog model, Deep Dyna-Q (DDQ). Furthermore, we designed learning schedules for SC-DDQ and DDQ, respectively, following two opposite training strategies: classic curriculum learning and its reverse version. 
		Our results show that by introducing scheduled learning and curiosity, the new framework leads to a significant improvement over the DDQ and Deep Q-learning(DQN). 
		Surprisingly, we found that traditional curriculum learning was not always effective. Specifically, according to the experimental results, the easy-first and difficult-first strategies are more suitable for SC-DDQ and DDQ. 
		To analyze our results, we adopted the entropy of sampled actions to depict action exploration and found that training strategies with high entropy in the first stage and low entropy in the last stage lead to better performance.
	\end{abstract}
	
	% keywords can be removed

	\section{Introduction}
	Since human-computer interaction and natural language processing are in high demand in industry and daily life, and the task-oriented dialog system has become a hot topic and deserves further study and research. 
	Dialog policy model is used to select the best action at each step of a dialog. In a task-oriented dialog system, the goal of the dialog is to complete a specific task, such as booking a movie. The dialog policy should select actions to efficiently achieve the goal.
	
	Dialog policy learning is often formulated as a reinforcement learning (RL) problem \cite{young2013pomdp,levin1997learning}, 
	in which a dialog agent executes an action based on the observed state and receives a reward from the environment acted by real users. 
	However, optimizing RL agents requires a large number of interactions with the environment, which is an expensive and time-consuming process. 
	Therefore, how to grasp dialog policy efficiently within limited interactions remains a fundamental question for RL \cite{lipton2018bbq,wu2019switch,zhang2020recent,mnih2015humanlevel,peng2018deep}.
	
	A common approach is to encourage the agent to explore the environment as sufficiently as possible through limited interactions. In contrast to supervised and unsupervised learning, reinforcement learning does not contain a large amount of training data with or without labels prepared in advance, but rather generates experience through the agent's interactions with the environment and continually optimizes the agent based on feedback from the environment. Therefore, exploration of the environment is crucial for RL, and extensive research has been conducted on promoting agents to explore the environment.
	Count-based exploration is an intuitive approach in which each state observed by the agent is recorded and the agent is encouraged to learn states with fewer occurrences. The most evident drawback of this approach is its inability to handle complex and high-dimensional data. To address this, Bellemare et al. engineered a model that acquires the distribution of visited environment states to measure the novelty of states and encourage agents to observe unfamiliar states \cite{bellemare2016unifying}. In addition, to deal with high-dimensional data, Tang et al. suggested mapping high-dimensional states to hash codes \cite{tang2017exploration}, and Liu et al. introduced embedding networks to encode the state space \cite{liu2022count}. However, a common flaw in such approaches is that they do not lead agents to explore states that have never existed before.
	
	Another method is to model the intrinsic motivation of the agents. One ingenious strategy is to utilize curiosity to facilitate state exploration. Pathak et al. \cite{pathak2017curiosity} proposed an Intrinsic Curiosity Module (ICM) to model the state prediction error, which depicts the uncertainty and improvement in RL as a curiosity reward to encourage unfamiliar dialog states. This research makes some simplifications and adjustments to ICM to make it more suitable for task-oriented dialog systems to guide unfamiliar dialog state exploration. This model leveraged the predictability of a state to depict its novelty. If the next state can be accurately predicted based on the current state and the upcoming action to be performed, it means that the action cannot lead the agent to see unfamiliar states, and it corresponds to a less curiosity reward. By contrast, if the next state is difficult to predict, the corresponding action receives a larger curiosity reward. The agent executes the action with the largest sum of curiosity rewards and rewards from the environment.
	
	Furthermore, the RL-based dialog agent faces another challenge.
	Generally, an episode of dialog based on RL is opened up by randomly selecting
	a user goal from the entire training dataset. Because the
	subsequent conversation revolves around the drawn goal until it is
	achieved successfully, it has an important influence on the
	subsequent training process. Random sampling neglects the way humans
	acquire knowledge and skills, where they focus on relatively easy
	materials before harder ones, and hurts the efficiency and stability of
	the training process. To overcome this problem, 
	Bengio et al. \cite{Bengio2009} proposed curriculum learning (CL), which presents relatively easy or simple examples at an early stage, inspired by human learning habits.
	Many NLP tasks have been improved by using a typical CL
	\cite{platanios2019competence,tay2019simple,xu2020curriculum,wang2020curriculum,liu2021scheduled,zhao2020reinforced,zhao2021automatic,zhu2021cold}. These learning methods employed
	different evaluations of training examples and approaches to adjust
	the training steps but shared the easy-first strategy.
	
	However, in some studies, the reverse version of CL was tested and
	achieved the best performance among various training scheduler designs
	\cite{zhao2020reinforced,chang2017active,hacohen2019on}. The
	effectiveness and application of easy-to-hard and hard-to-easy strategies are still worth exploring \cite{wang2021survey}. 
	Chang et al. \cite{chang2017active}
	stated that the difficult-first strategy is more suitable for
	cleaner datasets, \hl{whereas} the classic CL is beneficial for acquiring policies
	through noisy scenarios and leads to faster convergence. Furthermore,
	when tasks are difficult for \hl{an} agent to complete, \hl{earlier presentation of}
	easier samples is preferable for an effective training process.
	However, the exact impact of scheduled training on agent behavior and
	policy optimization remains unclear.
	
	This study aims to improve the performance of a task-oriented dialog system by providing sufficient environment exploration and a training strategy that matches human learning habits. Therefore, we propose Scheduled Curiosity-Deep Dyna-Q (SC-DDQ) which combines CL and a curiosity reward with dialog policy optimization based on Deep Dyna-Q (DDQ), where DDQ is a
	state-of-the-art model-based dialog system \cite{peng2018deep}, and design
	a curriculum based on two opposite learning strategies, which are found to
	be optimal in different scenarios. Compared to count-based exploration approaches \cite{bellemare2016unifying,tang2017exploration,liu2022count}, and variants of DDQ, SC-DDQ for the first time adapts the curiosity model which models the agent's intrinsic motivation, and introduces it into DDQ to motivate the agent to explore the environment based on its intrinsic motivation. In addition, SC-DDQ develops criteria for classifying the difficulty of training samples based on the capabilities of the dialog agent and introduces CL into the training of the dialog agent.
		
	Because the specific application scenarios of easy-to-difficult and difficult-to-easy learning strategies are not yet conclusive, in the experimental section, we carried out four combinations of experiments based on the presence or absence of the curiosity model and the trend of the difficulty of the task, aiming to explore the effects of agent intrinsic motivation and learning strategies on the completion of the dialog tasks. Preferring more difficult user goals is effective when experiments are conducted without curiosity reward while focusing on easier tasks is found to be an optimal training strategy if the dialog agent selects its action according to both external and internal rewards.
	
	To determine and analyze the influences of scheduled training on agent behavior, the entropy of agent action sampling in each training stage, measuring the agent action exploration, was employed to characterize the policy strategy. It is worth noting that entropy is designed to analyze the experimental data, not part of the algorithm itself. 
	We point out that a higher entropy at an earlier stage yields better performance. In other words, at the beginning of learning, relatively uniform action sampling,
	namely more sufficient action space exploration, leads to a higher task
	success rate. According to the experimental results, a key
	factor in facilitating dialog task completion was encouraging action
	exploration during the early training stage.
	
	Compared with \cite{lipton2018bbq,wu2019switch,zhang2020recent,bellemare2016unifying,tang2017exploration,liu2022count}, SC-DDQ is more scalable. First, this study makes some simplifications to ICM which is the original curiosity model. It performs feature extraction on the state to filter out parts of the state that are unaffected by action execution. Because almost all elements contained in the representation of the dialog state in the context of movie ticket booking scenarios are manipulated by action execution, feature extraction is discarded in this research, which should be borrowed for other scenarios whose states do not contain too many interfering parts. Second, the introduction of CL is mainly an adaptation of the training sequence; therefore, it can be easily extended to algorithms other than dialog systems. Furthermore, our experimental results provide evidence that different training scenarios can potentially benefit from both classical CL and its inverse version. This result presents new ideas for subsequent research on CL. Therefore, the SC-DDQ framework is scalable to real-world applications such as cyber defense. Moradi et al. leveraged reinforcement learning to acquire an attack strategy and allocate defense resources to protect smart electrical power grids\cite{moradi2022defending,moradi2023preferential}. The main idea of SC-DDQ can be adapted to these two cyber defense frameworks. For example, the curiosity reward can be modeled as the prediction error of the defense resource allocation status, and the order of the training tasks presented to the agent can be adjusted according to their defense difficulties.

	In summary, our main contributions in this paper are three-fold:
	
	\begin{enumerate}
		\def\labelenumi{(\arabic{enumi})}
		\item
		We propose a curiosity-driven scheduled framework extended on the Deep Dyna-Q (DDQ) model to improve the performance and learning efficiency of task-oriented dialog systems. To the best of our knowledge, this is the first adaptation of the curiosity model, integrating it into a task-oriented dialog system. This result is presented in Section \ref{the-effectiveness-of-the-proposed-SC-DDQ-and-its-variants}.
		\item
		We designed learning curricula based on opposite training strategies
		and presented both benefits to policy grasping under different reward
		settings. Compared with common applications of CL, this research is not limited to an intuitive easy-to-difficult training strategy, and for the first time, opposite training strategies are implemented in different settings of the same framework. This contribution is presented in Section \ref{the-effectiveness-of-two-opposite-training-strategies}. 
		\item
		We adopt the entropy of agent action sampling to depict
		the behavioral characteristics of the agent, and point out that guiding the agent to
		attempt various actions in the early phase of training facilitates
		policy optimization, which is explained in detail in Section \ref{observation-of-entropy-of-action-sampling}.
	\end{enumerate}
	
	\section{Related work}
	
	\subsection{Deep Dyna-Q}
	
\begin{figure}[H]
		\centering
		\begin{minipage}[t]{0.49\textwidth}
			\centering
			\includegraphics[width=6cm]{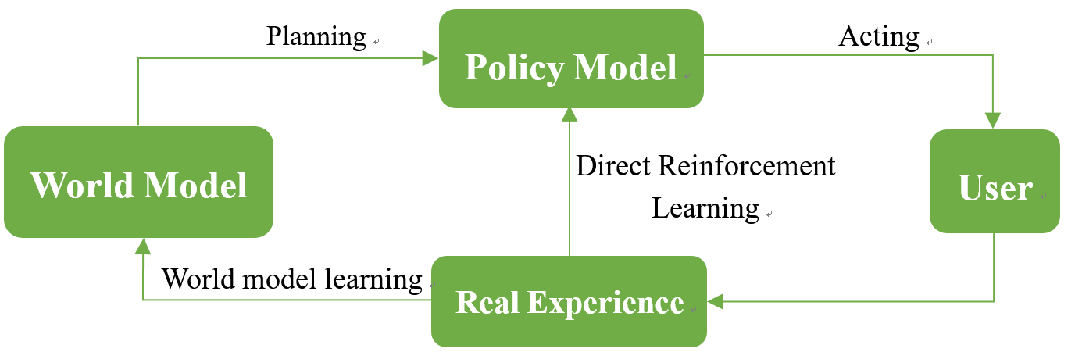}
			\caption{The framework of DDQ}
			\label{figure1}
		\end{minipage}
		\begin{minipage}[t]{0.49\textwidth}
			\centering
			\includegraphics[width=6cm]{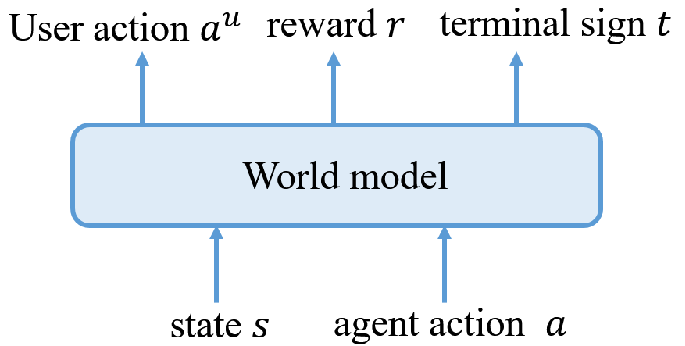}
			\caption{The structure of world model}
			\label{figure2}
		\end{minipage}
	\end{figure}

	Our research is based on the Deep Dyna-Q (DDQ) model, a classic model-based RL dialog system \hl{that integrates} planning to
	improve \hl{the} task completion rate within limited interactions \cite{peng2018deep}. 
	The framework is illustrated in Fig. \ref{figure1}. It consists of three processes: (1)
	direct reinforcement learning, \hl{in which} the policy model learns from
	real dialog experience; (2) planning, \hl{in which} the world model is applied
	to generate a simulated experience to improve the dialog policy
	model; \hl{and} (3) world model learning, \hl{in which} the world model is improved
	through real experience.

	\subsubsection{Direct Reinforcement Learning}
	Dialog policy learning can typically be formulated as a Markov Decision
	Process (MDP). An episode of task-oriented dialog can be regarded as a
	set of tuples. In each dialog turn, the policy agent observes the
	dialog state \(s\) and samples action \(a\). The selection of 
	action \(a\) is based on $\varepsilon$-greed policy, where the action is chosen
	randomly with probability $\varepsilon$ or to maximize the action-value
	function \(Q(s,a;\theta)\). The action-value function is accomplished by
	a Multi-Layer Perceptron (MLP) with parameters \(\theta_{Q}\). After
	executing this action, the agent receives a reward \(r\) from the
	environment and observes the user response \(a^{u}\). The dialog
	state is then updated to the next state \(s^\prime\). The tuple
	\((s,\ a,\ r,\ a^{u},\ s^\prime)\) can be viewed as a piece of experience
	and stored in either a real or a simulated experience replay buffer. The
	action-value function $Q(\cdot)$ is improved via Deep Q-network (DQN) \cite{mnih2015humanlevel}. 
	
	An episode of dialog is launched by sampling a user goal from a goal
	set by the user simulator. Each user goal is defined as G =
	(inform\_slots, request\_slots), where inform\_slots is a set of
	constraints and request\_slots is a set of requests. For the
	movie-ticket booking task, typical information slots  
	involve items such as movie names or number of people. Requests can be theater or start time.
	
	\subsubsection{Planning and World Model Learning}
	
	In the planning process, the world model, which simulates the environment, interacts with the policy agent and generates simulated experiences stored in a simulated replay buffer. In each planning turn,
	state \(s\) and agent action \(a\) are viewed as inputs to the
	world model. The world model then generates its response \(a^{u}\), the
	corresponding reward \(r\), and a binary variable \(t\) indicating
	whether this episode is over, as shown in Fig. \ref{figure2}. Both
	direct reinforcement learning and planning were accomplished using the same
	DQN algorithm, training on real and simulated experience. The world model was implemented as an MLP and tuned \hl{based} on real experiences during the world model learning process.
	
	\subsection{Curriculum Learning}
	
	Curriculum learning (CL) is a training strategy \hl{in which} the agent initially concentrates on relatively easy data and progressively moves to harder training samples. This strategy is inspired by human study characteristics and is beneficial for accelerating \hl{the} training convergence speed and enhancing \hl{the} agent performance. 
	CL is more implementable and compatible than multitudinous and complicated frameworks and models, \hl{thereby} making it more universally applicable.
	In addition, CL makes the best use of the existing data, 
	which \hl{saves} resources and time for RL-based tasks.
	
	Typical CL \hl{have} been \hl{used} extensively in natural language processing. Liu et al. \cite{liu2021scheduled} employed the mastering level of current tasks as the difficulty criterion and guided the agent to grasp easier tasks first.
	Zhao et al. \cite{zhao2021automatic} 
	utilized the total number of inform and request slots \hl{for} tasks to identify tasks, and the task completion rate to adjust the difficulty level \hl{of the next task}.
	
	However, the easy-to-hard learning strategy is not always helpful.
	In some studies, the opposite version of CL, focusing on harder
	examples, brings more significant improvements to the system. Zhao et al.
	\cite{zhao2020reinforced} proposed an opposite CL strategy, \hl{in which} the agent was trained on
	a general training set and gradually moved to subsets. Moreover,
	Hacohen et al. \cite{hacohen2019on} demonstrated that self-paced learning, a \hl{well-known} variant of CL, hurts the agent's performance. 
	Chang et al. \cite{chang2017active} stated that CL is more suitable for training scenarios with 
	\hl{significant} noise. \hl{By} contrast, the training process of cleaner scenarios
	would be more efficient using the opposite CL strategy.
	
	As the effectiveness and applicability of \hl{difficult-first and easy-first strategies} are still open to debate, we utilized the difficulty
	of user goals as a measurement of dialog complexity and proposed
	opposing learning schedules based on CL and its reverse version.
	Applying different reward functions yields a performance boost for the dialog agents based on the two learning schedules.
	
	\subsection{Curiosity Reward}
	Curiosity is modeled by state error prediction to promote the agent
	to attempt unconversant states \cite{schmidhuber1991possibility,pathak2017curiosity}.
	In RL applications \cite{li2019curiosity,wesselmann2019curiosity,Bougie2020,
		li2020random,wangananont2022simulation}, \hl{an} agent
	samples an action and receives a reward after observing the current
	state. \hl{Subsequently}, the \hl{entire} dialog environment updates to the next state. If
	\hl{a} new state can be predicted accurately 
	before selecting the following action, \hl{it} is viewed as a familiar state for the agent.
	\hl{However}, a state that is \hl{difficult} \hl{to} forecast is more informative for
	the agent. In this \hl{study}, we integrate the curiosity model with dialog
	policy optimization \hl{to guide} agents to explore unfamiliar environment
	states. 
	
	In this research, we simplified and adjusted the Intrinsic Curiosity Module (ICM) \cite{pathak2017curiosity}. ICM is engineered to play video games, such as VizDoom and Super Mario Bros, and it observes screenshots of the game screen as inputs. Because the game screen often contains many disturbances that are not controlled by the action, such as the change of scenery in the background, ICM designs an inverse model to filter out these disturbing factors and thus performs feature extraction on the states. Unlike the application scenarios of ICM, the states in this research are affected by the execution of actions and contain almost no interference factors. Therefore, the curiosity model used in this research removed the inverse model. In addition, ICM calculates the prediction error as the curiosity reward between the next state and the predicted one after the agent executes the action and observes the true next state. In this research, the agent does not execute the feasible actions individually to observe the corresponding next state and calculate the curiosity reward; instead, the agent takes both the curiosity reward and the predicted next state as outputs for training. This eliminates the need for the agent to repeatedly step back, but the disadvantage is that at the early stage of training, the curiosity reward output from our curiosity model may not be accurate enough.
	
	Previous studies introduced curiosity-based exploration from a video game scenario into an RL-based dialog system and achieve performance breakthroughs. Wang et al. \cite{wang2019dialogue} investigated different exploration strategies, including ICM, in task-completion dialog policy learning and showed improvements. Doering et al. \cite{doering2019curiosity} developed a shopkeeper robot whose verbal interactions with customers are guided by curiosity, and it is significantly human-like, compared to non-curious robots. These studies provide evidence for the potential effectiveness of the curiosity model in dialog systems. Therefore, we made some adjustments to the ICM, integrated it into the dialog system, and believe in its effectiveness.
	
	\section{Proposed methods}
	\subsection{Overview}
	
	The common training strategy for dialog agents is to randomly expose the agent to tasks with different difficulties, where the training efficiency is low and the agent's performance \hl{can} be hurt. In addition, the dialog environment acted by real users is too complicated to be fully explored. Therefore, we propose Scheduled Curiosity-Deep Dyna-Q (SC-DDQ), a novel and
	practical framework applying a curiosity strategy \hl{to} joint curriculum
	learning and RL-based policy learning for task-oriented dialog \hl{systems},
	implemented based on Deep Dyna-Q to improve the dialog agent performance.
	
	\begin{figure}[H]
		\begin{center}
			\includegraphics[width=0.6\columnwidth]{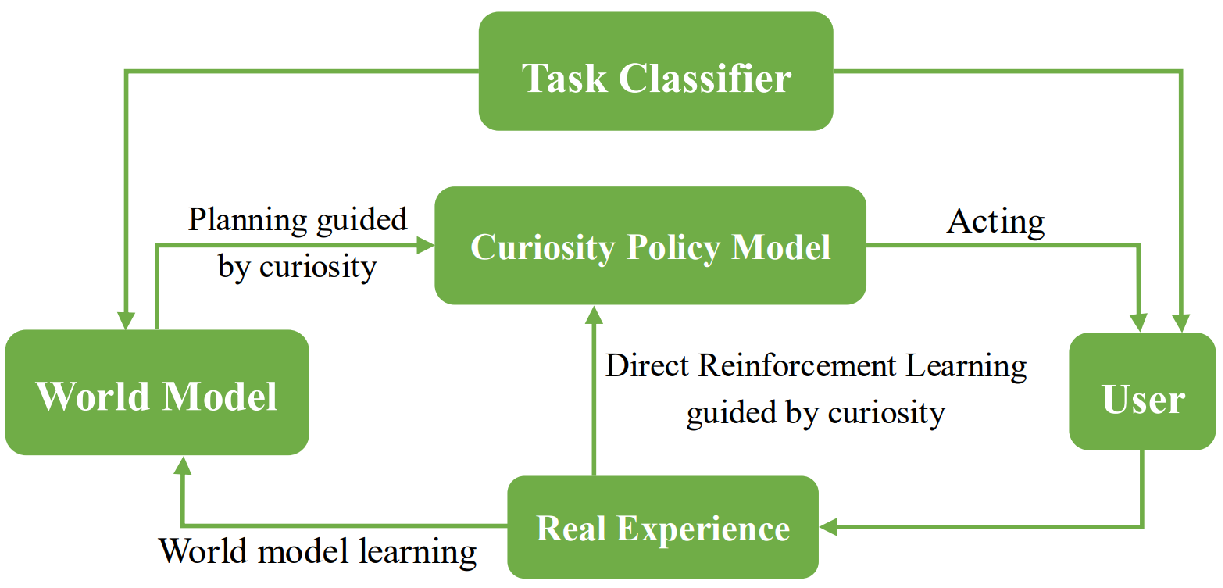}
		\end{center}
		\caption{The framework of Scheduled Curiosity-Deep Dyna-Q}
		\label{figure3}
	\end{figure}
	
	The proposed framework is illustrated in Fig. \ref{figure3}. The SC-DDQ framework consists of three modules: (1) an off-line task classifier for dividing user goals into three complexity levels; (2) a curiosity policy agent for selecting \hl{the} next action beneficial for completing the task and exploring state by using the current dialog state; \hl{and} (3) a world model for simulating a user to generate actions and rewards based on real user behaviors.
	
	The optimization of SC-DDQ comprises five steps: 
	(1) warm starting: 
	a hand-crafted dialog policy is employed to generate experiences;
	(2) direct reinforcement learning guided by curiosity: 
	the agent guided by the curiosity model interacts with real users \hl{and} generates real experiences. \hl{They are stored in the real experience replay buffer and used to improve dialog policy};
	(3) world model learning: the
	world model is refined on real experiences; 
	(4) planning guided by curiosity: the agent guided by the curiosity model conducts interactions with the world model, where the simulated experiences are collected and employed to optimize dialog policy;
	\hl{and} (5) curiosity model training: the curiosity model is optimized
	using both real and estimated experiences.
	
	\hl{In addition}, the proposed framework can be equipped with different policy
	models and curriculum schedules. Four combinations are
	implemented in this research, as shown in Table \ref{table1}: (1) policy agent without
	curiosity \hl{equipped} with \hl{the easy-first strategy (EFS) curriculum; (2) policy agent without
		curiosity equipped with the difficult-first strategy (DFS) curriculum; (3) curiosity-combined
		policy agent equipped with the EFS curriculum; and (4) curiosity-combined
		policy agent equipped with the DFS curriculum.}
	
	\begin{table}[H]
		\caption{Four combinations}
		\label{table1}
		\begin{center}
			\begin{tabular}{l|cc}
				\hline
				Method & Curiosity & Schedule \\
				\hline
				DDQ & No & Random \\
				C-DDQ & Yes & Random \\
				S-DDQ\_EFS & No & EFS \\
				S-DDQ\_DFS & No & DFS \\
				SC-DDQ\_EFS & Yes & EFS \\
				SC-DDQ\_DFS & Yes & DFS \\
				\hline
			\end{tabular}
		\end{center}
	\end{table}
	
	In this research, two opposite types of curriculum training schedules are proposed, both of \hl{which} enable the RL dialog agent to \hl{achieve} better performance under certain conditions. To reveal the influence of the training curriculum on the performance of the agent, the response of the agent in each turn of \hl{the} dialog \hl{was} recorded. These results are presented in Section 5.2.
	
	The structure of the main parts of SC-DDQ \hl{is described} in the subsequent subsection. 
	The pseudo-code \hl{for} the iterative SC-DDQ algorithm is \hl{shown} in Algorithm 1. \hl{Line 1 describes the first step in the optimization, where a rule-based dialog is conducted. Lines 2-12 show that the off-line task classifier divides user goals into three levels. Lines 14-17 illustrate that before launching an episode of dialog a user goal that fits the curriculum is sampled. The second step of optimization is described in lines 18-24, where the curiosity policy model outputs Q-values and curiosity rewards and selects the best feasible action based on their sum. Line 24 describes the update of the DQN based on real experiences. The third step of optimization corresponds to line 25 where the world model is updated using real experiences. Line 26 illustrates the fourth step, where the world model, replacing the user simulator, interacts with the agent and generates simulated experiences. Line 27 shows that the DQN is optimized by simulated experiences. The last step of the optimization is shown in line 28, where the curiosity model is updated using both real and simulated experiences.}
	\subsection{Task Classifier}
	\begin{figure}[H]
		\begin{center}\small\tabcolsep=1pt
			\begin{tabular}{|p{2cm}p{6cm}|}
				\hline
				\multicolumn{2}{|c|}{User goal: }\\
				\hline
				\multicolumn{2}{|l|}{\parbox{8cm}{
						\texttt{\footnotesize
							\begin{tabbing}
								XX \= XX \= \kill
								\{ \\
								\>  "request\_slots": \{ \\
								\>\>  "ticket": "UNK" \\
								\>  \}, \\
								\>  "diaact": "request", \\
								\>  "inform slots": \{  \\
								\>\>    "city": "Seattle", \\
								\>\>    "number of people": "2", \\
								\>\>    "theater": "amc pacific place 11 theater", \\
								\>\>    "start time": "10:00 pm",\\
								\>\>    "date": "tomorrow",\\
								\>\>    "movie name": "race"\\
								\>  \}\\
								\}
							\end{tabbing}
				}}}\\
				\hline
				\multicolumn{2}{|c|}{Dialog conducted by rule-based agent:}\\
				\hline
				Turn 0 usr: &I want tickets for race at 10:00 pm. \\
				Turn 1 sys: &What movie are you interested in? \\
				Turn 2 usr: &I want to watch race.\\
				Turn 3 sys: &What time would you like to see it? \\
				Turn 4 usr: &I want to watch at 10:00 pm.\\
				Turn 5 sys: &Which city would you like?\\
				Turn 6 usr: &I want to watch at Seattle.\\
				Turn 7 sys: &What date would you like to watch it? \\
				Turn 8 usr: &I want to set it up tomorrow\\
				Turn 9 sys: &Which theater would you like?\\
				Turn 10 usr: &I want to watch at amc pacific place 11 theater.\\
				Turn 11 sys: &How many tickets do you need?\\
				Turn 12 usr: &I want 2 tickets please!\\
				Turn 13 sys: &Great - I was able to purchase 2 tickets for you to see race tomorrow at amc pacific place 11 theater in Seattle at 10:00 pm. \\
				Turn 14 usr: &Thank you.\\ 
				Turn 15 sys: &Thank you.\\
				\hline
			\end{tabular}
		\end{center}
		\caption{A success task completed by rule-based agent}
		\label{figure4}
	\end{figure}
	
		\begin{table}[H]
		\caption{The data annotation schema}
		\label{table2}
		\begin{center}
			\begin{tabular}{l|p{6cm}}
				\hline
				Intent & request, inform, deny, confirm\_question, \\
				& confirm\_answer, greeting, closing, not\_sure, \\
				& multiple\_choice, thanks, welcome\\
				\hline
				Slot & city, closing, date, distanceconstraints, \\
				& greeting, moviename, numberofpeople, price, \\
				& starttime, state, taskcomplete, theater, \\
				& theater\_chain, ticket, video\_format, zip\\
				\hline
			\end{tabular}
		\end{center}
	\end{table}
	In this research, we designed an offline classifier focusing on booking movie tickets,
	identifying user goals
	into three complexity levels: easy, middle, and difficult. A user goal comprises two parts: request and inform slots. Inform slots are constraints known or \hl{determined} by the user. \hl{The} request slots are unknown to the user and must be obtained from the responses of the agent. Table \ref{table2} lists all slots. \hl{Consider} Fig. \ref{figure5} (b) \hl{as} an \hl{example}. This user goal contains three request slots: ticket, theater, and start time, and three inform slots: number of people, date, and movie name. This goal reveals that the user would like to buy three tickets for Zootopia tonight, but he/she has yet to decide which theater to go to and \hl{at a} specific time. During this episode of dialog, the agent should provide information about \hl{a} suitable theater and start time to satisfy the user's constraints.
	In addition, the ticket is a default slot that always appears in the request slots \hl{for} user goals.

	Before launching the iterative scheduled SC-DDQ, the system is
	opened up with a warm starting phase, where a rule-based dialog policy
	\cite{peng2018deep} is used to interact with users and generate
	experiences. 
	\hl{As} the dialog strategy of the rule-based agent requires the
	movie name, start time, city, date, theater, and number of people in
	order, a user goal with a limited number of request slots is a breeze for
	an agent to achieve, as a successful example shown in Fig. \ref{figure4}. 
	Therefore,
	this research applies the number of request slots as the measurement of
	the difficulty of training tasks, rather than the length of dialog,
	word rarity, and the total number of requests and inform slots
	\cite{sachan2016easy,platanios2019competence,see2019what,cai2020learning,zhao2020reinforced}. The goal classification \hl{was} performed offline. In this study, we classified user goals by the number of request slots\hl{, as listed in lines 2-12 of Algorithm 1.}
	
	Fig. \ref{figure5} shows \hl{additional} examples of the user goals at different levels. 
	As shown in
	Fig. \ref{figure5}(a), goals with only one request slot are classified as easy and stored in the easy goal buffer \(G_{\textnormal{easy}}\). Goals with two or three slots are then classified as middle and stored in the
	middle goal buffer \(G_{\textnormal{middle}}\). Finally, goals with four
	or five slots are classified as difficult and stored in the difficult goal
	buffer \(G_{\textnormal{difficult}}\). At the
	beginning of the dialog, the user samples a goal from the corresponding set. In this research, schedules designed on two opposite training strategies
	are conducted, which are Easy-Middle-Difficult (EMD),
	Easy-Difficult-Difficult (EDD), Easy-Easy-Difficult-All (EED),
	Difficult-Middle-Easy (DME), Difficult-Easy-Easy (DEE) and
	Difficult-Difficult-Middle
	(DDM). 
	
	\hl{Training on the SC-DDQ framework contained 300 epochs following the setting of the DDQ \cite{peng2018deep}. At the beginning of each epoch, the user simulator selects a goal from the goal buffer according to a specific schedule, as shown in lines 14 and 15 of Algorithm 1. Taking the schedule Difficult-Middle-Easy (DME) as an example, from epochs 0 to 69, the user simulator randomly chooses a goal from the difficult goal buffer, supposing that the sampled goal is the goal shown in Fig. \ref{figure5}(c). Then, the user simulator launches this episode of dialog with the first utterance, for example, \textit{when is Deadpool playing in Los Angeles?} Subsequently, the agent performs an action, such as informing the start time of the Deadpool. In turn, the user opens the next turn of the dialog, and it continues until the selected user goal is achieved, or the dialog is terminated by too many rounds. From epochs 70 to 139, the user simulator turned to sample goals from the middle goal buffer. The easy goal buffer was leveraged from epochs 140 to 209. Finally, from epochs 210 to 299, all user goals participate in the training, namely, the user simulator samples user goals from the total goal buffer}\(G_{\textnormal{total}}\).

	\begin{figure}[H]
		\begin{center}\footnotesize
			\begin{tabular}{c|c|c|c|c|c}
				\hline
				\multicolumn{6}{c}{\textbf{Easy goal}} \\
				\hline
				\multicolumn{3}{c|}{Request slots} & 
				\multicolumn{3}{c}{Inform slots} \\
				\hline
				1 & Ticket & UNK & 1 & City & Seattle\\
				&          &         & 2 & Number of people & 2\\
				&          &         & 3 & Theater &Royal theater \\
				&          &         & 4 & Start time &10:00 pm\\
				&          &         & 5 & Date &Tomorrow\\
				&          &         & 6 & Movie name&Race\\
				\hline
			\end{tabular} \\
			(a) An easy goal \\
			\medskip
			\begin{tabular}{c|c|c|c|c|c}
				\hline
				\multicolumn{6}{c}{\textbf{Middle goal}} \\
				\hline
				\multicolumn{3}{c|}{Request slots} & 
				\multicolumn{3}{c}{Inform slots} \\
				\hline
				1 & Ticket & UNK       & 1 & Number of people & 3 \\
				2 & Theater & UNK     & 2 & Date & Tonight \\
				3 & Start time & UNK  & 3 & Movie name & Zootopia \\
				\hline
			\end{tabular}\\
			(b) A middle goal \\
			\medskip
			\begin{tabular}{c|c|c|c|c|c}
				\hline
				\multicolumn{6}{c}{\textbf{Difficult goal}} \\
				\hline
				\multicolumn{3}{c|}{Request slots} & 
				\multicolumn{3}{c}{Inform slots} \\
				\hline
				1 & Ticket & UNK        & 1 & City & Los Angeles \\
				2 & Date & UNK          & 2 & Number of people& 1 \\
				3 & Theater & UNK      & 3 & Movie name & Deadpool \\
				4 & Start time & UNK   &   & & \\
				\hline
			\end{tabular}\\
			(c) A difficult goal \\
		\end{center}
		\caption{User goals in different difficulty level. UNK means that the corresponding slot is unknown.}
		\label{figure5}
	\end{figure}
	
	\subsection{Curiosity Policy Model}
	
\begin{figure}[H]
		\begin{center}
			\includegraphics[width=8cm]{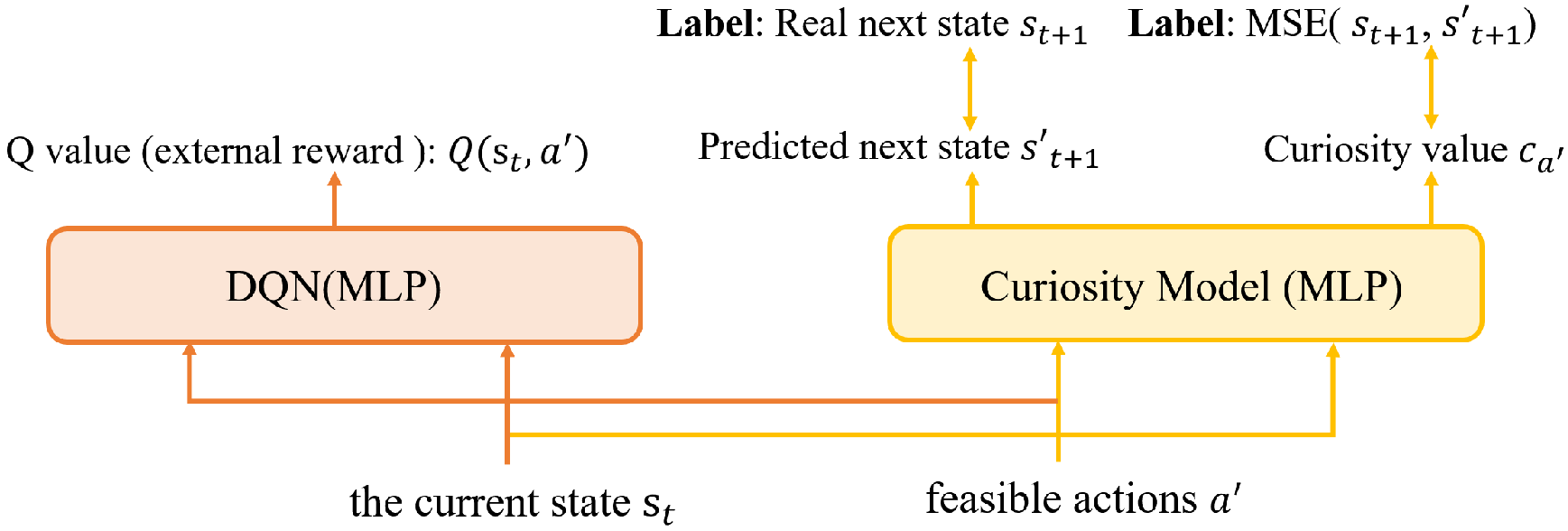}
		\end{center}
		\caption{The structure of the curiosity policy model}
		\label{figure6}
	\end{figure}

	The proposed curiosity policy model is a \hl{combination} of the DQN-based agent \cite{mnih2015humanlevel}
	and \hl{the} curiosity model, as shown in Fig. \ref{figure6}. The
	DQN-based agent \(Q(s_t,a_t;\theta_{Q})\) is used to interact with users,
	and the curiosity model \(C(s_t,a_t;\theta_{C})\) is used to generate
	\hl{a} curiosity value to assist the agent \hl{in} selecting an action \hl{that} leads to
	an unfamiliar state.
	
	\textbf{During direct reinforcement learning}, in step $t$, the agent observes state \(s_{t}\) and chooses an action \(a_{t}\) to carry into the next dialog turn. Traditionally, the action \(a_{t}\) is chosen according to $\epsilon$ -greedy, 
	where we choose a random action with probability $\epsilon$, otherwise the action is chosen following the greedy policy \(a_{t}\ = \text{argmax}_{a^\prime} Q\left( s_{t},\ a^\prime;\theta_{Q}\right)\). 
	The $a^\prime$ represents feasible agent actions, and \(Q\left( s,\ a;\theta_{Q}\right)\) is the approximated value function, implemented as a Multi-Layer Perceptron (MLP) parameterized by $\theta_{Q}$. To encourage \hl{the exploration of} unfamiliar states, we introduce \hl{the} curiosity value generated by the curiosity model in this step. The curiosity model was implemented \hl{using} MLP \(C\left( s,\ a;\theta_{C}\right)\). As shown in Fig. \ref{figure6}, this model takes the current state \(s_{t}\) and candidate actions $a^\prime$ as inputs, \hl{and} the predicted next state \(s_{t+1}^\prime\) and the curiosity value \(c_{a^\prime}\) as outputs. A larger curiosity value indicates a larger difference between the real and predicted next state. For an agent, a state that cannot be predicted accurately is unfamiliar and informative. Therefore, the action maximizing the addition of the Q value and curiosity value \hl{is} sampled, which can be formulated as Equation 1: 
	\begin{equation}
		a_{t} = argmax_{a^\prime}\left[ Q\left( s_{t},\ a^\prime;\theta_{Q} \right) + c_{a^\prime}\right]
		\label{eq3}
	\end{equation}
	\hl{Subsequently}, the agent receives a reward \(r_{t}\) from the environment, observes the next user action \(a^{u}\), and updates to the next state \(s_{t+1}\) until \hl{it reaches} the end of \hl{the} dialog. Experience \((s_{t},\ a_{t},\ r_{t},\ a^{u},\ s_{t + 1})\) is stored in the real replay buffer \(D^{u}\). We improve the Q value function \(Q\left( s,\ a;\theta_{Q}\right)\) using real experiences from \(D^{u}\) via minibatch SGD. \hl{It should be noted that the weight of the curiosity value cannot be adjusted for the time being. In other words, curiosity-based exploration cannot be  adjusted adaptively in this study.}
	
	\textbf{In the planning and world model learning process}, as described in
	the related work section, the world model is applied to interact with the agent and generate simulated experiences, which can also be used to improve the dialog policy. 
	The planning process was similar to \hl{that of} direct reinforcement learning. In this process, the environment is acted by the world model accomplished by an MLP \(M\left( s,\ a;\theta_{M}\right)\) as shown in Fig. \ref{figure2}. In each planning turn, the world model takes the current state \(s_{t}\) and the agent action \(a_{t}\) as inputs. The outputs are user action \(a^{u}\), reward \(r_{t}\), and a binary variable \(t\) indicating whether this episode is over or not. Dialog experiences generated during planning are stored in the simulated replay buffer \(D^{u}\). 
	The proposed framework follows the DDQ structure \cite{peng2018deep}, which has two
	experience replay buffers $D^u$ and $D^s$ for
	storing real and simulated
	experiences, respectively. 
	\hl{Because} the role of the world model is \hl{to imitate} real users, real experiences from \(D^{u}\) \hl{were} employed for its optimization via minibatch SGD.
	
	\textbf{In the curiosity model training process}, the curiosity model
	accomplished by MLP \(C(s,\ a,\ \theta_{C})\) is refined via minibatch
	SGD using experiences from both real and simulated replay buffers, namely
	\(D^{u}\) and \(D^{s}\). Its inputs are the current state \(s_{t}\) and candidate actions \(a^\prime\), \hl{whereas} its outputs are the predicted next state \(s_{t+1}^\prime\) and curiosity value \(c_{a^\prime}\). \hl{The} labels of the predicted next state \(s_{t+1}^\prime\) and the curiosity value \(c_{a^\prime}\) are the real next state \(s_{t+1}\) and state prediction error, respectively. To define the error of state prediction, we encode real and predicted next states \(s_{t+1}\) and \(s_{t+1}^\prime\) into vectors $\phi(s_{t+1})$ and $\phi(s_{t+1}^\prime)$ accomplished by one-hot encoding. The state prediction error is expressed as \( \ \left\| \phi\left(s_{t + 1}\right) - \ \phi\left(s_{t + 1} ^\prime\right) \right\|^{2}\).
	
	\begin{algorithm}[H]
		\SetAlgoLined
		\KwIn{Dialog user goal set 
			$G_\text{\scriptsize total}=\{g_1,\ldots,g_{n_\text{\tiny total\_user\_goals}}\}$, $N$}
		\KwResult{$Q(s,a;\theta_Q),M(s,a;\theta_M),C(s,a;\theta_C)$}
		The rule-based agent generates real experiences and stores them into $D^u$\;
		\ForAll{$level\in\{$\text{easy, middle, difficult}$\}$}{
			$G_{level}\gets\emptyset$\;
		}
		\For{$i \gets 1$ \KwTo $n_\textrm{\scriptsize total\_user\_goals}$}{
			\Switch{$N_\text{\scriptsize request\_slot}(g_i)$}{
				\lCase{\rm 1}{$level\gets$ easy}
				\lCase{\rm 2,3}{$level\gets$ middle}
				\lOther{$level\gets$ difficult}
			}
			Add $g_i$ into $G_{level}$\;
		}
		\For{$n\gets 1$ \KwTo $N$} {
			Determine $level$\;
			Sample a goal $g$ from $G_{level}$\;
			Set the initial state $s_0$ and the user's first action $a_0$\;
			$t\gets 1$\;
			\While{\rm dialog not terminated}{
				DQN {$Q(s,a;\theta_Q)$} generates Q-value\;
				Curiosity model {$C(s,a;\theta_C)$} generates curiosity value {$c_{a^\prime}$} and the predicted next state \(s_{t+1}^\prime\)\;
				Agent executes action $a=\arg\max_{a^\prime}[Q(s_t,a^\prime;\theta_Q)+c_{a^\prime}]$,
				observes the next user action $a^{u}$, receives reward $r_t$ and updates to the next dialog state
				$s_{t+1}$ \;
				Stores the real experience $(s_t, a_t, r_t, a^{u}, s_{t+1})$ to $D^u$\;
			}
			
			Update {$\theta_{Q}$} using real experiences from {$D^{u}$} via minibatch SGD\;
			Update {$\theta_{M}$} via minibatch SGD on real experiences from {$D^{u}$}\;
			World model interacts with the curiosity policy agent and generates simulated experiences stored to {$D^{s}$}\;
			Update {$\theta_{Q}$} using simulated experiences from {$D^{s}$} via minibatch SGD\;
			Update {$\theta_{C}$} using real and simulated experiences via minibatch SGD\;
		}
		\caption{Scheduled Curiosity-Deep Dyna-Q policy learning}
		\label{algorithm1}
	\end{algorithm}
	
	In addition, in each training stage, the actions selected by the agent
	are counted, and sampled action distributions are generated for each agent. The entropy of the sampled agent actions \hl{was} calculated to reveal the impact of the difficulty level of tasks on the behavior of the agent. \hl{It is worth emphasizing that entropy is employed only for the analysis of experimental results and is not part of the algorithm itself.}
	
	\section{Experimental Setup}
	
	\subsection{Datasets}
	
	The raw conversational data in movie-tickets booking dataset \cite{li2018microsoft}
	\footnote{https://github.com/xiul-msr/e2e\_dialog\_challenge} is
	collected via Amazon Mechanical Turk. The dataset was manually labeled based on a schema established by domain experts. The annotation schema
	has 11 intents and 16 slots, as shown in Table \ref{table2}. The original dataset
	contained up to four request slots for user goals. To thoroughly explore the
	effects of \hl{the} curriculum on dialog policy, we supplemented nine user goals containing five
	request slots. \hl{Thus, in total, 137 user goals were pre-generated for the movie-ticket
		booking scenario.} As shown in Table \ref{table3}, there were 61 goals
	with one request slot, 16 goals with two request slots, 17 goals with three
	request slots, 34 goals with four request slots, and nine goals with five request
	slots. Goals with only one request slot are easy. Then, goals with two or three slots were classified as middle. Finally, goals with four or five slots were classified as difficult. \hl{A total of 991 movies were available.} The information for each movie contains the movie name, date, start time, theater, etc.. There are 29 feasible actions for the agent and 35 feasible actions for the real user and the world model.
	
	\begin{table}[H]
		\caption{User goals with different number of request\_slot.
			$N_\textrm{\scriptsize request\_slot}$ is the number of request slots,
			$N_\textrm{\scriptsize goal}$ is the number of corresponding goals, and
			$N_\textrm{\scriptsize gset}$ is the total number of goals in the corresponding set.}
		\label{table3}
		\begin{center}
			\begin{tabular}{l|c|c|c}
				\hline
				Goals set& $N_\textrm{\scriptsize request\_slot}$ &$N_\textrm{\scriptsize goal}$  &$N_\textrm{\scriptsize gset}$ \\
				\hline
				Easy & 1 & 61 & 61 \\
				\hline
				Middle& 2 & 16 & 33 \\
				& 3 & 17 & \\
				\hline
				Difficult& 4 & 34 & 43 \\
				& 5 & 9 & \\
				\hline
			\end{tabular}
		\end{center}
	\end{table}
	
	\subsection{User Simulator}
	
	In this research, an accessible user simulator \cite{li2017user} was
	applied to evaluate the SC-DDQ model to \hl{ensure that} the training process
	was affordable and practical. During training, the user simulator offers
	rewards and simulated user responses according to hand-crafted rules to
	the dialog agent. An episode of dialog is judged \hl{to be} successful when \hl{the}
	agent provides a suitable movie ticket \hl{that satisfies} all \hl{constraints} from
	the user. The reward rules adopted by the user simulator are
	as follows: (1) in each turn, a reward of $-1$ is fed back to encourage a
	shorter dialog; (2) at the end of the dialog, a reward of $2L$ is provided
	for success or a reward of $-L$ is provided for failure, where $L$ is the
	maximum number of turns in each episode and is set to 40 in this
	research.
	
	\subsection{Training Schedules}
	
	Typical CL suggests presenting relatively easier samples before
	harder ones, namely the \hl{easy-first strategy (EFS)}. This study employs both
	the classic CL and its reverse version (difficult-first strategy, DFS)
	and designs several schedules:
	EMD, EDD, EED, DME, DEE, and DDM. Following the original DDQ setting, the \hl{entire} training process contains 300 epochs \cite{peng2018deep}, \hl{that were divided into four stages as evenly as possible}. The first to third stages, each consisting of 70 epochs, use the
	corresponding goals of the schedule condition. For example, in the EMD
	schedule, easy goals are used from 0 to 69 epochs, middle goals
	from 70 to 139 and difficult goals from 140 to 209 epochs. In the
	last stage, 210-299 epochs, the goals were uniformly sampled from all
	goals.
	
	\subsection{Examined Methods}
	
	We analyzed the impacts of the \hl{{blue} easy-first strategy (EFS) and difficult-first strategy (DFS)} by comparing several task-oriented dialog agents that employed variations of Algorithm 1.
	
	\begin{description}
		\item[DQN]: A task-completion dialog agent learned by standard DQN,
		implemented by direct reinforcement learning without \hl{a} curiosity model \cite{mnih2015humanlevel}.
		
		\item[DDQ]: A state-of-\hl{the-}art task-oriented dialog agent as described in
		the related work section \cite{peng2018deep}. 
		
		\item[S-DDQ]: A DDQ agent trained following specific schedules.
		
		\item[C-DDQ]: A DDQ agent equipped with a curiosity model but without
		curriculum learning.
		
		\item[SC-DDQ]: A DDQ agent equipped with a curiosity model and trained following specific schedules.
	\end{description}
	
	\subsection{Implementation Details}
	
	The DQN \(Q(s,a;\theta_{Q})\), curiosity model \(C(s,a;\theta_{C})\),
	and world model \(M(s,a;\theta_{M})\) were all accomplished using MLP
	with tanh activations. Following the original DDQ settings \cite{peng2018deep}, 
	the DQN contains one hidden layer with 80 hidden nodes.
	The world model had two shared hidden layers and three task-specific
	hidden layers, each \hl{with} 80 nodes. 
	The curiosity model contained
	two hidden layers and two task-specific hidden layers, each \hl{with} 80 nodes.  \(\varepsilon\)-greedy algorithm was adopted for exploration. Each
	network was optimized using RMSProp. The batch size was 16, and the discount
	factor was 0.9. The sizes of the real and simulated replay buffers,
	\(D^{u}\) and \(D^{s}\), were set to 5000. The target network is updated
	at the end of each training epoch. The maximum length of the simulated
	dialog was 40. When a dialog exceeded the turn limit, it was judged as a failure.
	In addition, to increase training efficiency, we utilized Reply Buffer
	Spiking (RBS) \cite{lipton2018bbq} and pre-filled the real experience
	replay buffer \(D^{u}\) \hl{in} the initial training stage with a set of real dialog experiences generated by a rule-based agent \cite{peng2018deep}.
	
	\section{Evaluation}\label{evaluation}
	The task success rates and \hl{the} average number of turns are presented in
	Tables \ref{table4} and \ref{table5}, respectively. 
	Each run was tested on 50 episodes of dialogs. The user goals
	for testing were sampled from the corresponding goals of the schedule
	conditions. Three main results are presented in the following subsections.
	\begin{table}[H]
		\caption{Success rate results. The stage 1 (S1), stage 2 (S2), stage 3
			(S3), and stage 4 (S4) shows the results at epoch 70, 140, 210, and 300.}
		\label{table4}
		\centering
		\begin{center}
			\begin{tabular}{lll|rrrr}
				\hline
				Strategy&Method&Schedule&\multicolumn{1}{c}{S1}&\multicolumn{1}{c}{S2}&\multicolumn{1}{c}{S3}&\multicolumn{1}{c}{S4}\tabularnewline
				\hline
				Random&DQN&-&$0.38$&$0.58$&$0.66$&$0.74$\tabularnewline
				Random&DDQ&-&$0.64$&$0.60$&$0.88$&$0.86$\tabularnewline
				Random&C-DDQ&-&$0.62$&$0.88$&$0.90$&$0.90$\tabularnewline
				\hline
				EFS&S-DDQ&EMD&$1.00$&$0.58$&$0.00$&$0.74$\tabularnewline
				EFS&S-DDQ&EDD&$1.00$&$0.00$&$0.56$&$0.88$\tabularnewline
				EFS&S-DDQ&EED&$1.00$&$1.00$&$0.00$&$0.62$\tabularnewline
				EFS&SC-DDQ&EMD&$0.98$&$0.58$&$0.64$&$\textbf{0.92}$\tabularnewline
				EFS&SC-DDQ&EDD&$0.98$&$0.54$&$0.66$&$0.88$\tabularnewline
				EFS&SC-DDQ&EED&$0.98$&$1.00$&$0.66$&$0.88$\tabularnewline
				\hline
				DFS&S-DDQ&DME&$0.56$&$0.74$&$1.00$&$0.88$\tabularnewline
				DFS&S-DDQ&DEE&$0.56$&$1.00$&$1.00$&$\textbf{0.94}$\tabularnewline
				DFS&S-DDQ &DDM&$0.56$&$0.70$&$0.82$&$\textbf{0.94}$\tabularnewline
				DFS&SC-DDQ&DME&$0.00$&$0.10$&$1.00$&$0.56$\tabularnewline
				DFS&SC-DDQ&DEE&$0.00$&$1.00$&$1.00$&$0.60$\tabularnewline
				DFS&SC-DDQ&DDM&$0.00$&$0.00$&$0.40$&$0.78$\tabularnewline
				\hline
		\end{tabular}\end{center}
	\end{table}
	
	\begin{table}[H]
		\caption{Average turn results. The stage 1 (S1), stage 2 (S2), stage 3
			(S3), and stage 4 (S4) shows the results at epoch 70, 140, 210, and 300.}
		\label{table5}
		\begin{center}
			\begin{tabular}{lll|rrrr}
				\hline
				Strategy&Method&Schedule&\multicolumn{1}{c}{S1}&\multicolumn{1}{c}{S2}&\multicolumn{1}{c}{S3}&\multicolumn{1}{c}{S4}\tabularnewline
				\hline
				Random&DQN&-&$31.28$&$23.36$&$25.48$&$23.32$\tabularnewline
				Random&DDQ&-&$23.12$&$22.36$&$15.44$&$18.20$\tabularnewline
				Random&C-DDQ&-&$24.57$&$22.73$&$25.32$&$16.68$\tabularnewline
				\hline
				EFS&S-DDQ&EMD&$24.48$&$28.00$&$39.09$&$24.88$\tabularnewline
				EFS&S-DDQ&EDD&$24.48$&$37.40$&$29.12$&$19.56$\tabularnewline
				EFS&S-DDQ&EED&$24.48$&$17.53$&$39.72$&$24.56$\tabularnewline
				EFS&SC-DDQ&EMD&$17.52$&$23.72$&$18.12$&$\textbf{13.36}$\tabularnewline
				EFS&SC-DDQ&EDD&$17.52$&$31.72$&$21.44$&$15.64$\tabularnewline
				EFS&SC-DDQ&EED&$17.52$&$16.73$&$22.88$&$22.20$\tabularnewline
				\hline
				DFS&S-DDQ &DME&$23.04$&$17.52$&$16.81$&$18.88$\tabularnewline
				DFS&S-DDQ &DEE&$23.04$&$19.08$&$17.48$&$16.36$\tabularnewline
				DFS&S-DDQ &DDM&$23.04$&$18.84$&$19.56$&$17.48$\tabularnewline
				DFS&SC-DDQ&DME&$35.16$&$36.54$&$17.28$&$26.16$\tabularnewline
				DFS&SC-DDQ&DEE&$35.16$&$16.68$&$13.61$&$24.27$\tabularnewline
				DFS&SC-DDQ&DDM&$35.16$&$41.93$&$35.12$&$31.40$\tabularnewline
				\hline
		\end{tabular}\end{center}
	\end{table}
	
	\subsection{The Effectiveness of the Proposed SC-DDQ and Its Variants}\label{the-effectiveness-of-the-proposed-SC-DDQ-and-its-variants}
\begin{figure}[H]
		\centering
		\begin{minipage}[t]{0.48\textwidth}
			\centering
			\includegraphics[width=8cm]{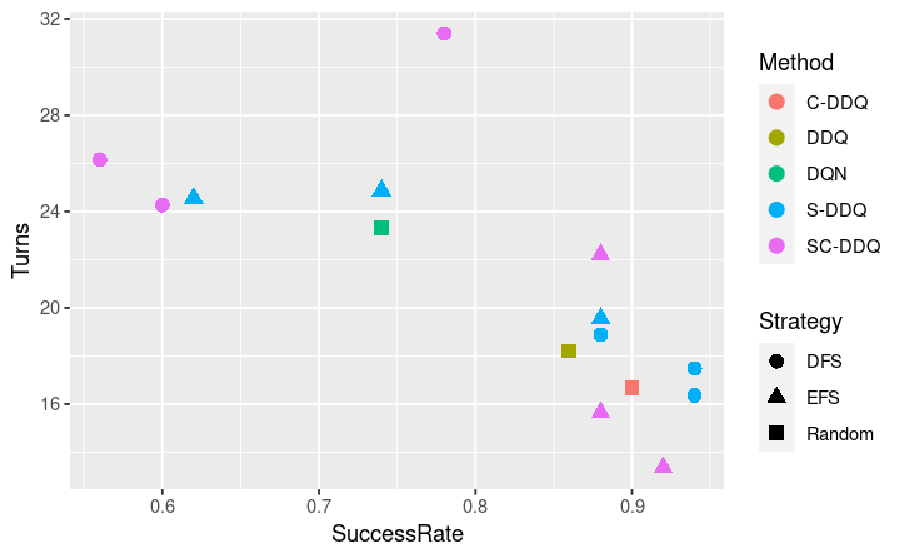}
			\caption{Success rates and average turns.}
			\label{figure11}
		\end{minipage}
		\begin{minipage}[t]{0.48\textwidth}
			\centering
			\includegraphics[width=9cm]{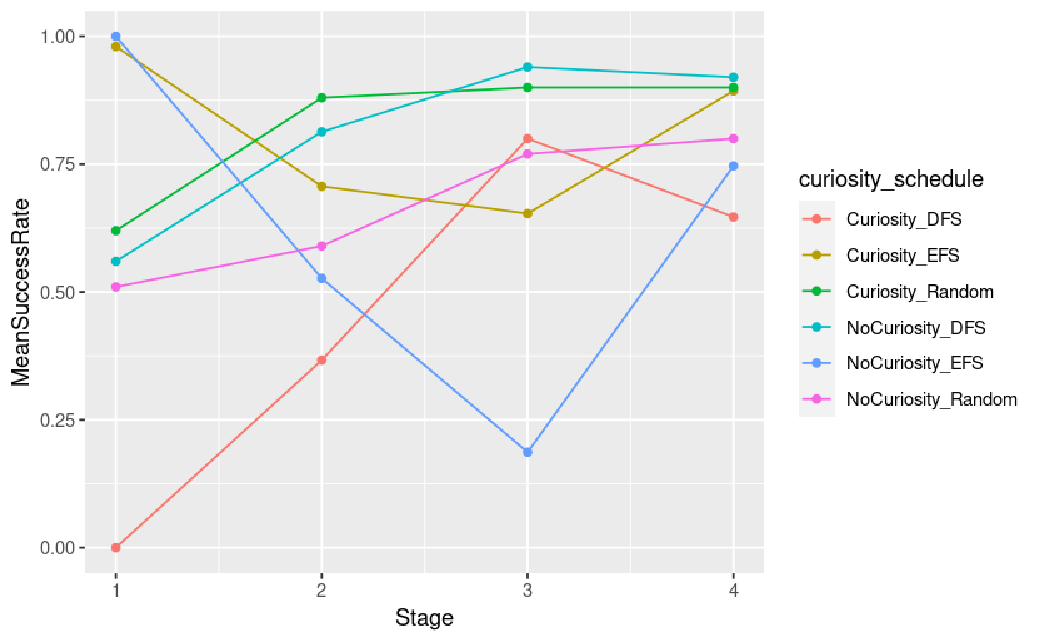}
			\caption{Success rates at each stage.}
			\label{figure12}
		\end{minipage}
	\end{figure}

	According to the experimental results, SC-DDQ and S-DDQ achieved outstanding performance compared with classic dialog agents, such as DQN and DDQ. As illustrated in Table \ref{table4}, schedule EMD yielded the highest success rate on SC-DDQ (0.92) at the final training stage, \hl{whereas} schedules DEE and DDM provided the best performance for S-DDQ (0.94), \hl{outperforming} DQN (0.74) and DDQ (0.86) by a large margin. Fig. \ref{figure11} shows the relationship between the success rate and the number of turns. This result clearly shows that a high success rate and small number of turns were achieved with better training
	strategies. \hl{In particular}, good results were obtained with \hl{DDQ using the difficult-first strategy (DFS) without curiosity (S-DDQ), or DDQ using the easy-first strategy (EFS) with curiosity (SC-DDQ).}
	S-DDQ with DFS yielded the best success rate, \hl{whereas} SC-DDQ with EFS
	achieved the smallest number of turns.
	
	\subsection{The Effectiveness of Two Opposite Training Strategies}\label{the-effectiveness-of-two-opposite-training-strategies}
	
	Both typical CL and its reverse version (difficult-first strategy, DFS) benefit from the dialog policy without or with curiosity, respectively. Fig. \ref{figure12} \hl{shows} the success rates for each training stage. The results for the same condition were averaged. For example, the
	\hl{curiosity\_DFS line} is the average of the SC-DDQ with DME,
	DDE, and DDM schedules. Overall, schedules designed by the typical CL (EFS) work
	better on agents with the curiosity policy, and DFS is beneficial for
	\hl{those} without the curiosity policy.
	
	Because the training of agents started with RBS, where the same rule-based
	policy designed for dealing with easy tasks was adopted, agents exposed to easy tasks initially showed a rapid performance boost, and the corresponding success rate reached 1.00. As the difficulty level of subsequent tasks increased, success rates
	decreased. Specifically, when the model without \hl{the} curiosity policy
	attempted to grasp difficult samples after \hl{exposure} to easy ones, its
	success rate decreased drastically, leading to poor training results
	in the final phase. For the model with \hl{the} curiosity policy, following EFS,
	its success rate decreased slightly in the middle of training; however, it
	picked up considerably towards the end of training thanks to the
	encouragement of state exploration from the curiosity model.
	\begin{figure}[H]
		\begin{center}\small\tabcolsep=1pt
			
			\begin{tabular}{|p{2cm}p{6.2cm}|}
				\hline
				\multicolumn{2}{|c|}{User goal: }\\
				\hline
				\multicolumn{2}{|l|}{\parbox{8cm}{
						\texttt{\footnotesize
							\begin{tabbing}
								XX \= XX \= \kill
								\{ \\
								\>"request\_slots": \{\\
								\>\>"ticket": "UNK"\\
								\>\},\\
								\>"diaact": "request",\\
								\>"inform slots": \{\\
								\>\>"city": "Birmingham", \\
								\>\>"number of people": "4",\\
								\>\>"theater": "Carmike summit 16", \\
								\>\>"state": "al",\\
								\>\>"start time": "around 6pm",\\t
								\>\>"date": "today",\\
								\>\>"movie name": "deadpool"\\
								\>\}\\
								\}
							\end{tabbing}
				}}}\\
				\hline
				\multicolumn{2}{|c|}{Dialog conducted by rule-based agent:}\\
				\hline
				Turn 0 usr:& Can I get 4 tickets for deadpool? \\
				Turn 1 sys:& Which city would you like? \\
				Turn 2 usr:& I want to watch at Birmingham. \\
				Turn 3 sys:& Carmike summit 16 is available.\\
				Turn 4 usr:& I want to watch at al.\\
				Turn 5 sys:& What time would you like to see it?\\
				Turn 6 usr:& I want to watch at around 6pm.\\
				Turn 7 sys:& What date would you like to watch it?\\
				Turn 8 usr:& I want to set it up today.\\
				Turn 9 sys:& Great, I was able to purchase 4 tickets to deadpool for today at around 6pm at Carmike summit 16 in Birmingham, al.\\
				Turn 10 usr:& Thank you.\\
				\hline
			\end{tabular}\\
			(a) An easy user goal\\
			\medskip
			\begin{tabular}{|p{2cm}p{6.2cm}|}
				\hline
				\multicolumn{2}{|c|}{User goal: }\\
				\hline
				\multicolumn{2}{|l|}{\parbox{8cm}{
						\texttt{\footnotesize
							\begin{tabbing}
								XX \= XX \= \kill
								\{ \\
								\>"request\_slots": \{\\
								\>\>"date": "UNK",\\
								\>\>"theater": "UNK",\\
								\>\>"start time": "UNK",\\
								\>\>"ticket": "UNK"\\
								\>\},\\
								\>"diaact": "request",\\
								\>"inform slots": \{\\
								\>\>"city": "Portland",\\
								\>\>"number of people": "4",\\
								\>\>"state": "Oregon",\\
								\>\>"movie name": "star wars"\\
								\>\}\\
								\}
							\end{tabbing}
				}}}\\
				\hline
				\multicolumn{2}{|c|}{Dialog conducted by rule-based agent:}\\
				\hline
				Turn 0 usr:& When is star wars playing in Oregon?\\
				Turn 1 sys:& Portland is available.\\
				Turn 2 usr:& What date is available?\\
				Turn 3 sys:& 12:05pm is available.\\
				Turn 4 usr:& What date is available?\\
				Turn 5 sys:& Friday is available.\\
				Turn 6 usr:& Which theater is available?\\
				Turn 7 sys:& Regal Lloyd center 10 \& IMAX is available.\\
				Turn 8 usr:& I want 4 tickets please!\\
				Turn 9 sys:& Great, I was able to purchase 4 tickets to star wars for Friday at 12:05pm at regal Lloyd center 10 \& IMAX in Portland, Oregon.\\
				Turn 10 usr:& Thank you.\\
				\hline
			\end{tabular}\\
			(b) A difficult user goal
		\end{center}
		\caption{Dialogs with different user goals}
		\label{figure7}
	\end{figure}
	When the agent first focused on difficult samples, the success rate at the early stage was lower than \hl{that} of EFS. 
	However, as the difficulty of the task decreased and the policy was
	improved, the success rates increased. In the final stage, \hl{the} success rates
	of both agents optimized according to DFS \hl{decreased,} except for schedule DDM.

		\begin{figure}[H]
		\begin{center}
			\includegraphics[width=7cm]{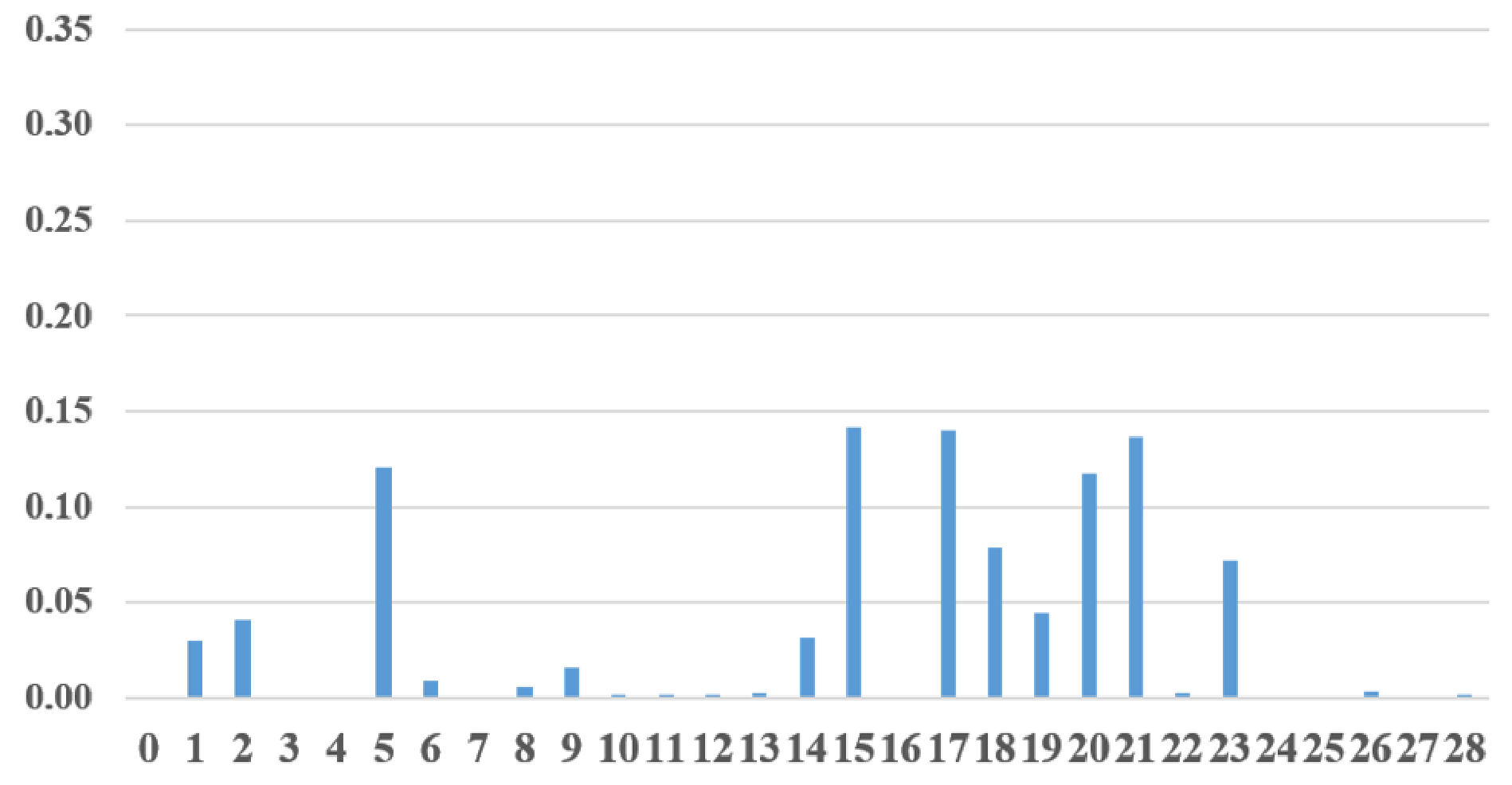}\\
			(a) Stage 1: Easy tasks\\
			\includegraphics[width=7cm]{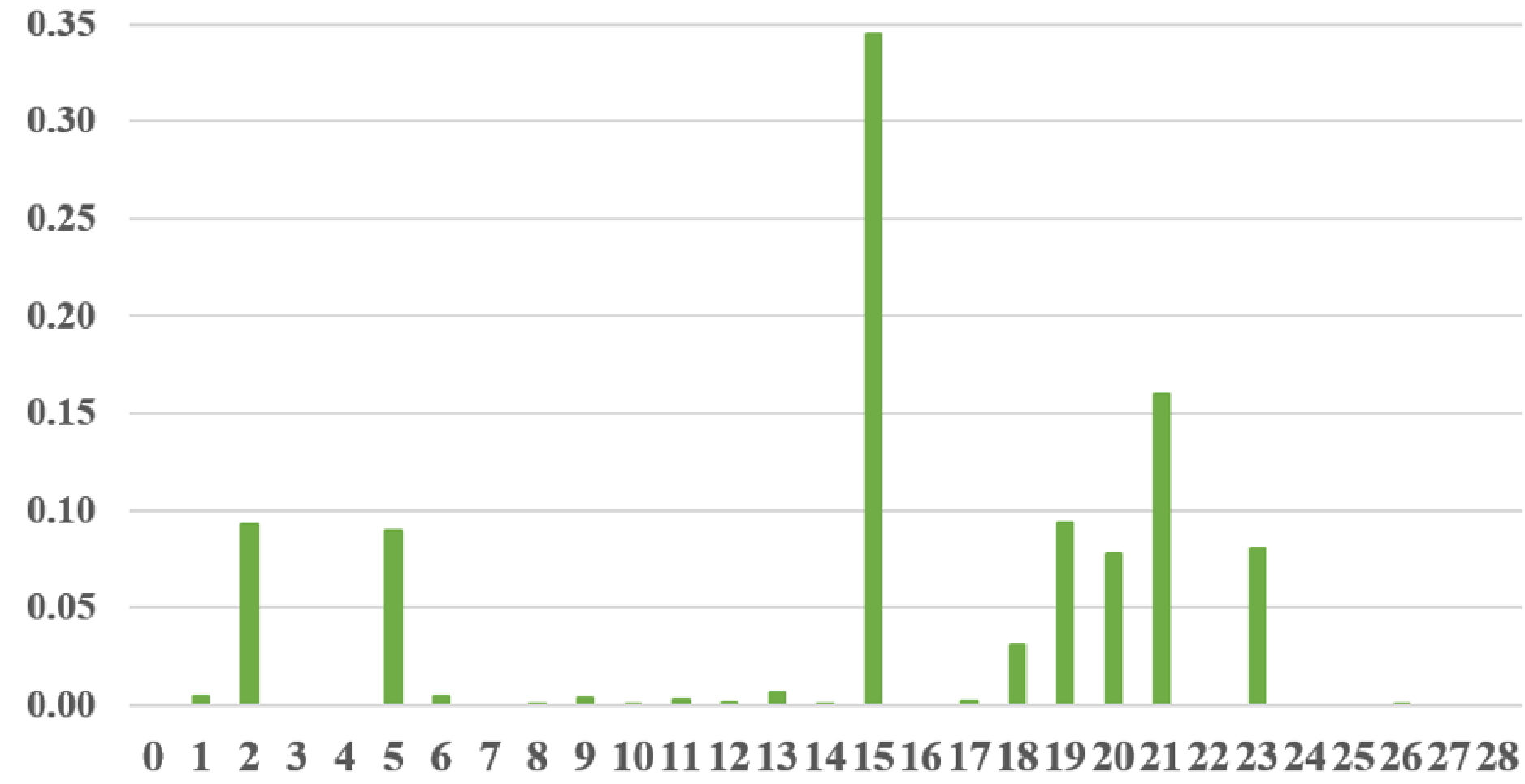}\\
			(b) Stage 2: Easy tasks\\
			\includegraphics[width=7cm]{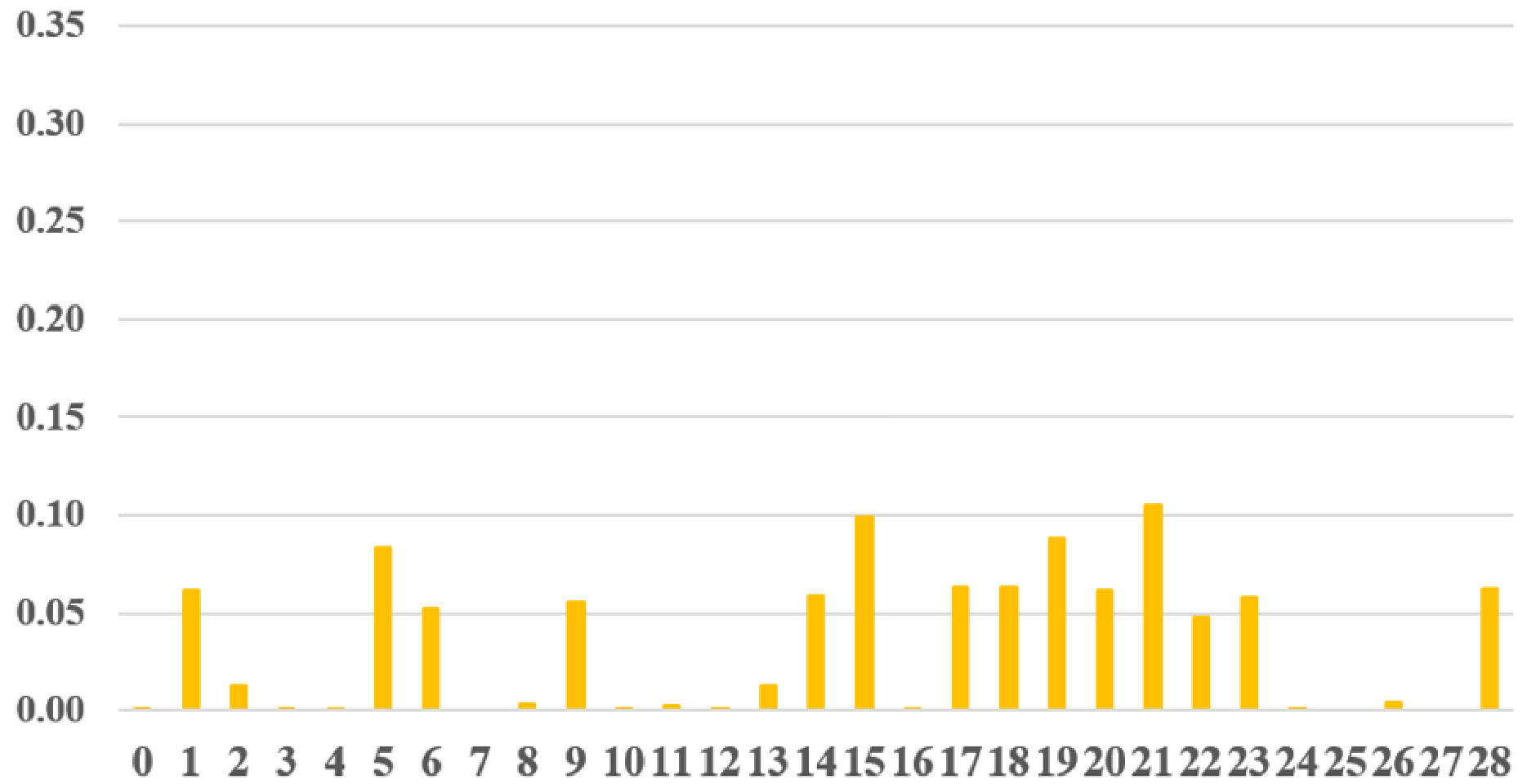}\\
			(c) Stage 3: Difficult tasks\\
			\includegraphics[width=7cm]{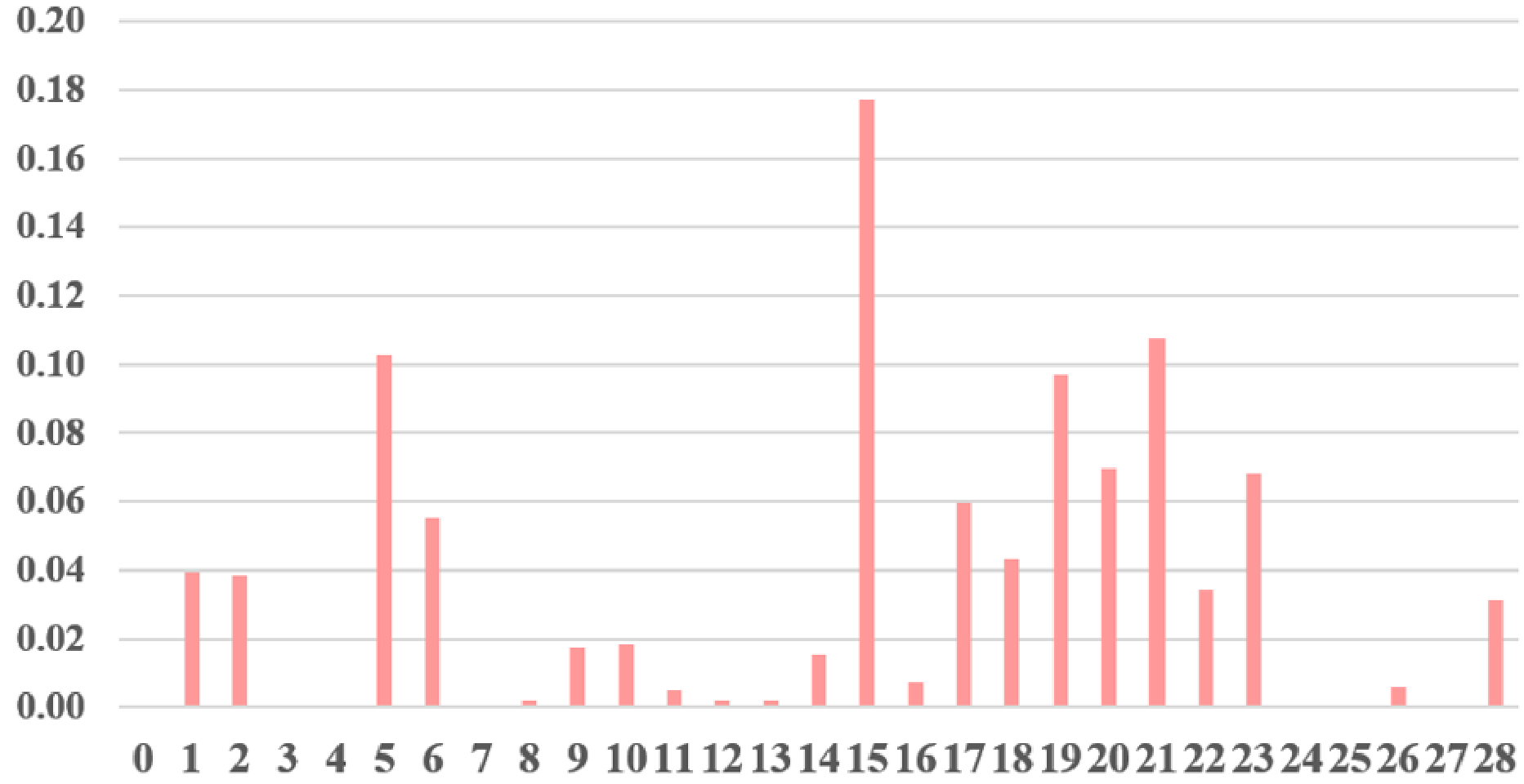}\\
			(d) Stage 4: All tasks\\
		\end{center}
		\caption{Distributions of actions sampled by S-DDQ EED.}
		\label{figure8} 
	\end{figure}
	\subsection{Observation of Entropy of Action
		Sampling}\label{observation-of-entropy-of-action-sampling}
	From the results shown in the previous section, we obtain\hl{ed} a high success rate \hl{under} two conditions: S-DDQ with DFS and SC-DDQ with EFS. To investigate why these conditions gave good results and other conditions did not, we introduced the entropy of sampled actions, which is used to depict the behavioral characteristics of agents. Agent sampling action preferences and \hl{entropy calculations} are explained in the following paragraph.
	
	When training at different task levels, the agent prefers to sample
	different actions. \hl{when} facing easy tasks, as shown in Fig. \ref{figure7} (a), the
	agent tends to ask more questions instead of providing information.
	\hl{When} facing difficult ones, as shown in Fig. \ref{figure7} (b), the actions sampled
	by the agent tend to \hl{diversify}. In other words, the agent
	attempts to provide appropriate information to answer the
	user query instead of \hl{simply} asking the user questions.
	
	Furthermore, the order in which the agent meets tasks of varying
	difficulty affects action sampling. Therefore, to explore the
	effect of training schedules on agent performance, we counted the actions
	employed by the agent in each training phase and calculated \hl{the frequency with which}
	each action was sampled. For example, Fig. \ref{figure8} (a), (b), (c), and (d)
	show the frequency of actions sampled by the agent without curiosity
	equipped with schedule EED. On schedule EED, the agent is
	sequentially trained with easy, easy, and difficult tasks for 70 epochs
	per training stage. Finally, it is trained with all tasks for 90
	epochs. The vertical coordinate is the frequency and the horizontal
	coordinate \hl{represents} the 29 possible actions. The entropy of the distribution
	of sampled actions is formulated as
	\begin{equation}
		H = - \sum_{i = 0}^{28}{P\left( a_{i} \right)\log{P(a_{i})}}
	\end{equation}
	\hl{Here,} $a_i$ represents feasible actions.
	$P(a_{i})$ is the probability of choosing action
	$a_{i}$ during the current training stage. Entropy
	represents the degree of uniformity \hl{in} sampling. A larger entropy \hl{implies that}
	the agent samples actions more evenly and attempts to perform various
	actions rather than being trapped in a few pseudo-optimal actions.

	Table \ref{table6} \hl{lists} the entropy values at each stage for all training
	conditions. Roughly speaking, in the S-DDQ with EFS conditions, the entropy values were small in the first stage and gradually \hl{increased} in the following stages. The entropy values of S-DDQ with DFS showed an opposite trend. \hl{In addition}, SC-DDQ with EFS showed large entropy in the first stage, \hl{whereas} that with DFS showed a random up-down trend.
	
	\begin{table}[H]
		\caption{Entropy values of each stage for all training conditions.}
		\label{table6}
		\begin{center}
			\begin{tabular}{lll|rrrr}
				\hline
				Strategy&Method&Schedule&\multicolumn{1}{c}{S1}&\multicolumn{1}{c}{S2}&\multicolumn{1}{c}{S3}&\multicolumn{1}{c}{S4}\tabularnewline
				\hline
				Random & DDQ & - & 3.98 & 4.16 & 3.76 & 3.63 \\
				Random & C-DDQ & - & 4.15 & 3.61 & 3.55 & 3.6 \\
				\hline
				EFS & S-DDQ & EMD & 3.53 & 3.58 & 3.97 & 3.93 \\
				EFS & S-DDQ & EDD & 3.53 & 3.68 & 3.87 & 3.55 \\
				EFS & S-DDQ & EED & 3.53 & 2.88 & 3.96 & 3.84 \\
				EFS & SC-DDQ & EMD & 4.04 & 3.77 & 4 & 3.76 \\
				EFS & SC-DDQ & EDD & 4.04 & 4.08 & 3.94 & 3.96 \\
				EFS & SC-DDQ & EED & 4.04 & 3.29 & 4.05 & 3.87 \\
				\hline
				DFS & S-DDQ & DME & 4.17 & 4.06 & 3.67 & 3.77 \\
				DFS & S-DDQ & DEE & 4.17 & 3.43 & 3.32 & 3.87 \\
				DFS & S-DDQ & DDM & 4.17 & 3.98 & 3.69 & 3.57 \\
				DFS & SC-DDQ & DME & 4.05 & 4.1 & 3.68 & 4.42 \\
				DFS & SC-DDQ & DEE & 4.05 & 3.45 & 3.37 & 4.23 \\
				DFS & SC-DDQ & DDM & 4.05 & 4.19 & 4.28 & 4.08 \\
				\hline
			\end{tabular}
		\end{center}
	\end{table}
	
	\section{Discussion}\label{discussion}
	
	Unlike traditional CL, in the present study\hl{,} we observed that EFS and DFS
	had different effects on task completion ability in different
	task-oriented dialog models. DFS is beneficial for S-DDQ, \hl{whereas}
	the classic CL is helpful for SC-DDQ. These two combinations share
	high entropy in the first stage and low entropy in the last stage,
	where sufficient action exploration leads to better performance.
	Contrary to intuition, employing a curiosity model designed to
	explore states is not always effective \hl{for} exploring actions.
	
\begin{figure}[H]
		\centering
		\includegraphics[width=8cm]{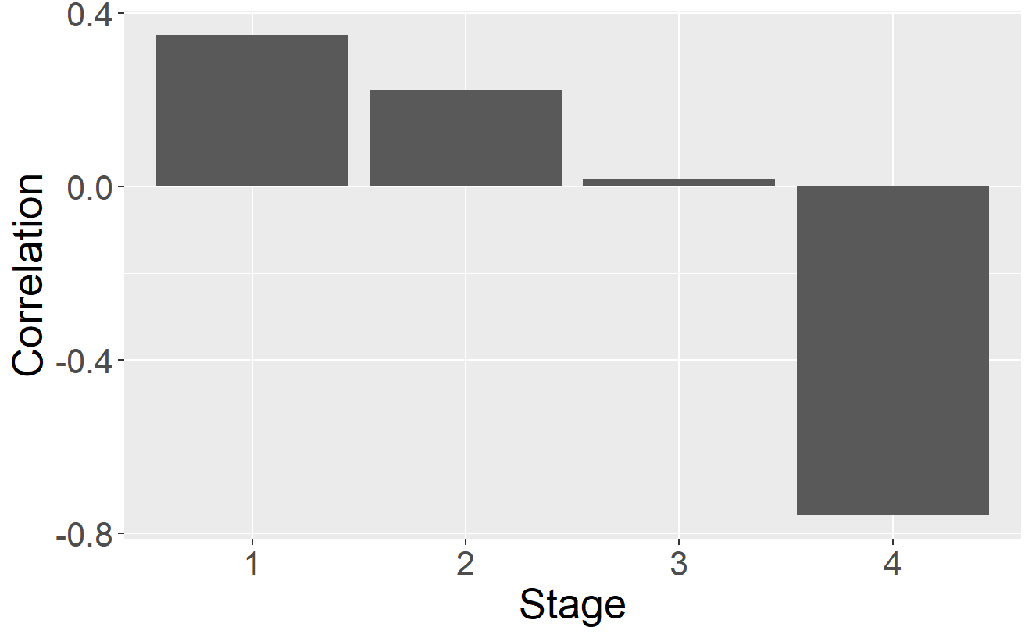}
		\caption{Correlation between entropy values at each stage and the
			success rate at the final stage.}
		\label{figure13}
	\end{figure}

	To investigate \hl{the effects of} the final success rate, we calculated
	Pearson's correlation coefficient between the entropy values at each
	stage and the final success rate. The results are shown in Fig. \ref{figure13}. This
	result shows that entropy in the first stage positively correlates
	with the success rate, \hl{whereas} entropy in the final stage is negatively
	correlated with the success rate. \hl{This} means that an agent with a high
	success rate samples the actions more randomly first, and the sampling
	diversity converges in the last stage. If \hl{curiosity is not employed},
	training with easy tasks leads to \hl{a greater} sampling bias and smaller
	entropy. This situation can be avoided using difficult tasks or curiosity. However, if we use difficult tasks at an early
	stage when using curiosity, the action sampling model does not converge and the success rate does not improve.
	
	\section{Limitations and Future works}
	\hl{This research has several limitations. First, the training schedules offered in this research were manually designed. Such schedules lack flexibility, and once designed, changes can not be made during the training process. Second, the curiosity reward cannot be adaptively adjusted for the time being. It participates in agent optimization with the same weight from beginning to end. However, dynamic adjustment of curiosity is necessary for its application in different scenarios. Finally, in the early training stage, the curiosity reward corresponding to each feasible action may deviate from its actual value. Compared with ICM \cite{pathak2017curiosity}, our curiosity model outputs the curiosity reward directly before updating to the next state, which allows the agent to obtain curiosity rewards before executing the action. Because the agent does not actually execute feasible actions individually and observe the next state, it does not obtain the real next state before acquiring the curiosity reward, and the curiosity reward output by the curiosity model is a simulation.}
	
	\hl{By analyzing the entropy of action sampling, we found that the dialog system exhibits better performance when the entropy tends to decrease. In other words, we advocate that the RL-based dialog agent could benefit from encouraging exploration in the early training stage and then gradually decreasing the level of exploration. Based on our findings, we plan to add weight to the curiosity reward to make curiosity-driven exploration more adaptive. Further work on the gradual discarding of curiosity is worth conducting, because this is a generic promising path for other RL problems.}
	\section{Conclusion}\label{conclusion}
	
	This paper presents a new framework, Scheduled Curiosity-DDQ (SC-DDQ), for
	task-oriented dialog policy learning. With the introduction of a curiosity
	model and scheduled training strategy, SC-DDQ outperforms previous
	classic dialog agents, such as DDQ and DQN. Moreover, variants of SC-DDQ
	\hl{were} conducted to verify the effectiveness and influence of a typical CL
	and its reverse version. Based on the experimental results, \hl{the difficult-first strategy benefits S-DDQ, whereas the easy-first strategy
		is preferable for SC-DDQ.} To explore the impact of task difficulty on the
	dialog agent policy, we calculated entropy. We found common
	trends of entropy, where the agent tried various actions randomly in the
	first stage, and converged in the last stage.
	
	\hl{In the future, we plan to control the curiosity-driven exploration during the training process based on entropy trends.} Furthermore, we believe that this curriculum
	framework can be applied to improve other RL-based
	NLP models.
	
	\bibliographystyle{unsrt}

	%\bibliography{reference}  %%% Uncomment this line and comment out the ``thebibliography'' section below to use the external .bib file (using bibtex) .


\begin{thebibliography}{10}
		
		\bibitem{young2013pomdp}
		M.~Young, S.and~Gasic, B.~Thomson, and J.~D. Williams.
		\newblock Pomdp-based statistical spoken dialog systems: A review.
		\newblock {\em Proceedings of the IEEE}, 101(5):1160--1179, 2013.
		
		\bibitem{levin1997learning}
		E.~Levin, R.~Pieraccini, and W.~Eckert.
		\newblock Learning dialogue strategies within the markov decision process
		framework.
		\newblock In {\em 1997 IEEE Workshop on Automatic Speech Recognition and
			Understanding Proceedings}, 1997.
		
		\bibitem{lipton2018bbq}
		Zachary Lipton, Xiujun Li, Jianfeng Gao, Lihong Li, Faisal Ahmed, and Li~Deng.
		\newblock Bbq-networks: Efficient exploration in deep reinforcement learning
		for task-oriented dialogue systems.
		\newblock In {\em Proceedings of the AAAI Conference on Artificial
			Intelligence}, volume~32, 2018.
		
		\bibitem{wu2019switch}
		Y.~Wu, X.~Li, J.~Liu, J.~Gao, and Y.-N. Yang.
		\newblock Switch-based active deep dyna-q: Efficient adaptive planning for
		task-completion dialogue policy learning.
		\newblock In {\em Proceedings of the AAAI Conference on Artificial
			Intelligence}, volume~33, pages 7289--7296, 2019.
		
		\bibitem{zhang2020recent}
		Z.~Zhang, R.~Takanobu, Q.~Zhu, M.~Huang, and X.~Zhu.
		\newblock Recent advances and challenges in task-oriented dialog systems.
		\newblock {\em Science China Technological Sciences}, 63(10):2011--2027, 2020.
		
		\bibitem{mnih2015humanlevel}
		V.~Mnih, K.~Kavukcuoglu, D.~Silver, A.~A. Rusu, J.~Veness, M.~G. Bellemare,
		A.~Graves, M.~Riedmiller, A.~K. Fidjeland, G.~Ostrovski, S.~Petersen,
		C.~Beattie, A.~Sadik, I.~Antonoglou, H.~King, D.~Kumaran, D.~Wierstra,
		S.~Legg, and D.~Hassabis.
		\newblock Human-level control through deep reinforcement learning.
		\newblock {\em Nature}, 518(7540):529--533, 2015.
		
		\bibitem{peng2018deep}
		B.~Peng, X.~Li, J.~Gao, J.~Liu, and K.~Wong.
		\newblock Deep dyna-q: Integrating planning for task-completion dialogue policy
		learning.
		\newblock In {\em Proceedings of the 56th Annual Meeting of the Association for
			Computational Linguistics (Volume 1: Long Papers)}, 2018.
		
		\bibitem{bellemare2016unifying}
		Marc Bellemare, Sriram Srinivasan, Georg Ostrovski, Tom Schaul, David Saxton,
		and Remi Munos.
		\newblock Unifying count-based exploration and intrinsic motivation.
		\newblock {\em Advances in neural information processing systems}, 29, 2016.
		
		\bibitem{tang2017exploration}
		Haoran Tang, Rein Houthooft, Davis Foote, Adam Stooke, OpenAI Xi~Chen, Yan
		Duan, John Schulman, Filip DeTurck, and Pieter Abbeel.
		\newblock \# exploration: A study of count-based exploration for deep
		reinforcement learning.
		\newblock {\em Advances in neural information processing systems}, 30, 2017.
		
		\bibitem{liu2022count}
		Xinyue Liu, Qinghua Li, Yuangang Li, et~al.
		\newblock Count-based exploration via embedded state space for deep
		reinforcement learning.
		\newblock {\em Wireless Communications and Mobile Computing}, 2022, 2022.
		
		\bibitem{pathak2017curiosity}
		D.~Pathak, P.~Agrawal, A.~A. Efros, and T.~Darrell.
		\newblock Curiosity-driven exploration by self-supervised prediction.
		\newblock In {\em 2017 IEEE Conference on Computer Vision and Pattern
			Recognition Workshops (CVPRW)}, 2017.
		
		\bibitem{Bengio2009}
		Y.~Bengio, J.~Louradour, R.~Collobert, and J.~Weston.
		\newblock Curriculum learning.
		\newblock In {\em Proceedings of the 26th Annual International Conference on
			Machine Learning}, 2009.
		
		\bibitem{platanios2019competence}
		E.~A. Platanios, O.~Stretcu, G.~Neubig, B.~Poczos, and T.~Mitchell.
		\newblock Competence-based curriculum learning for neural machine translation.
		\newblock In {\em Proceedings of the 2019 Conference of the North}, 2019.
		
		\bibitem{tay2019simple}
		Y.~Tay, S.~Wang, L.~A. Tuan, J.~Fu, M.~C. Phan, X.~Yuan, J.~Rao, S.~C. Hui, and
		A.~Zhang.
		\newblock Simple and effective curriculum pointer-generator networks for
		reading comprehension over long narratives.
		\newblock In {\em Proceedings of the 57th Annual Meeting of the Association for
			Computational Linguistics}, 2019.
		
		\bibitem{xu2020curriculum}
		B.~Xu, L.~Zhang, Z.~Mao, Q.~Wang, H.~Xie, and Y.~Zhang.
		\newblock Curriculum learning for natural language understanding.
		\newblock In {\em Proceedings of the 58th Annual Meeting of the Association for
			Computational Linguistics}, 2020.
		
		\bibitem{wang2020curriculum}
		C.~Wang, Y.~Wu, S.~Liu, M.~Zhou, and Z.~Yang.
		\newblock Curriculum pre-training for end-to-end speech translation.
		\newblock In {\em Proceedings of the 58th Annual Meeting of the Association for
			Computational Linguistics}, 2020.
		
		\bibitem{liu2021scheduled}
		S.~Liu, J.~Zhang, K.~He, W.~Xu, and J.~Zhou.
		\newblock Scheduled dialog policy learning: An automatic curriculum learning
		framework for task-oriented dialog system.
		\newblock In {\em Findings of the Association for Computational Linguistics:
			ACL-IJCNLP 2021}, 2021.
		
		\bibitem{zhao2020reinforced}
		M.~Zhao, H.~Wu, D.~Niu, and X.~Wang.
		\newblock Reinforced curriculum learning on pre-trained neural machine
		translation models.
		\newblock In {\em Proceedings of the AAAI Conference on Artificial
			Intelligence}, volume~34, pages 9652--9659, 2020.
		
		\bibitem{zhao2021automatic}
		Y.~Zhao, Z.~Wang, and Z.~Huang.
		\newblock Automatic curriculum learning with over-repetition penalty for
		dialogue policy learning.
		\newblock In {\em Proceedings of the AAAI Conference on Artificial
			Intelligence}, volume~35, pages 14540--14548, 2021.
		
		\bibitem{zhu2021cold}
		Hui Zhu, Yangyang Zhao, and Hua Qin.
		\newblock Cold-started curriculum learning for task-oriented dialogue policy.
		\newblock In {\em 2021 IEEE International Conference on e-Business Engineering
			(ICEBE)}, pages 100--105, 2021.
		
		\bibitem{chang2017active}
		H.~Chang, E.~Learned-Miller, and A.~McCallum.
		\newblock Active bias: Training more accurate neural networks by emphasizing
		high variance samples.
		\newblock In {\em Advances in Neural Information Processing Systems},
		volume~30, 2017.
		
		\bibitem{hacohen2019on}
		G.~Hacohen and D.~Weinshall.
		\newblock On the power of curriculum learning in training deep networks.
		\newblock In {\em Proceedings of the 36th International Conference on Machine
			Learning (PMLR)}, volume~97, pages 2535--2544, 2019.
		
		\bibitem{wang2021survey}
		X.~Wang, Y.~Chen, and W.~Zhu.
		\newblock A survey on curriculum learning.
		\newblock {\em IEEE Transactions on Pattern Analysis and Machine Intelligence},
		pages 1--1, 2021.
		
		\bibitem{moradi2022defending}
		Mohammadamin Moradi, Yang Weng, and Ying-Cheng Lai.
		\newblock Defending smart electrical power grids against cyberattacks with deep
		q-learning.
		\newblock {\em P R X Energy}, 1:033005, 2022.
		
		\bibitem{moradi2023preferential}
		Mohammadamin Moradi, Yang Weng, John Dirkman, and Ying-Cheng Lai.
		\newblock Preferential cyber defense for power grids.
		\newblock {\em P R X Energy}, 2:043007, Oct 2023.
		
		\bibitem{schmidhuber1991possibility}
		J.~Schmidhuber.
		\newblock A possibility for implementing curiosity and boredom in
		model-building neural controllers.
		\newblock In {\em From Animals to Animats}, 1991.
		
		\bibitem{li2019curiosity}
		B.~Li, T.~Lu, J.~Li, N.~Lu, Y.~Cai, and S.~Wang.
		\newblock Curiosity-driven exploration for off-policy reinforcement learning
		methods.
		\newblock In {\em 2019 IEEE International Conference on Robotics and
			Biomimetics (ROBIO)}, 2019.
		
		\bibitem{wesselmann2019curiosity}
		P.~Wesselmann, Y.-C. Wu, and M.~Gasic.
		\newblock Curiosity-driven reinforcement learning for dialogue management.
		\newblock In {\em ICASSP 2019 - 2019 IEEE International Conference on
			Acoustics, Speech and Signal Processing (ICASSP)}, 2019.
		
		\bibitem{Bougie2020}
		N.~Bougie and R.~Ichise.
		\newblock Fast and slow curiosity for high-level exploration in reinforcement
		learning.
		\newblock {\em Applied Intelligence}, 51(2):1086--1107, 2020.
		
		\bibitem{li2020random}
		J.~Li, X.~Shi, J.~Li, X.~Zhang, and J.~Wang.
		\newblock Random curiosity-driven exploration in deep reinforcement learning.
		\newblock {\em Neurocomputing}, 418:139--147, 2020.
		
		\bibitem{wangananont2022simulation}
		R.~Wangananont, N.~Buppodom, S.~Chanthanuraks, and V.~Kotrajaras.
		\newblock Simulation of homogeneous fish schools in the presence of food and
		predators using reinforcement learning.
		\newblock In {\em 2022 17th International Joint Symposium on Artificial
			Intelligence and Natural Language Processing (iSAI-NLP)}, 2022.
		
		\bibitem{wang2019dialogue}
		Yu-An Wang and Yun-Nung Chen.
		\newblock Dialogue environments are different from games: Investigating
		variants of deep q-networks for dialogue policy.
		\newblock In {\em 2019 IEEE Automatic Speech Recognition and Understanding
			Workshop (ASRU)}, pages 1070--1076, 2019.
		
		\bibitem{doering2019curiosity}
		Malcolm Doering, Phoebe Liu, Dylan~F Glas, Takayuki Kanda, Dana Kuli{\'c}, and
		Hiroshi Ishiguro.
		\newblock Curiosity did not kill the robot: A curiosity-based learning system
		for a shopkeeper robot.
		\newblock {\em ACM Transactions on Human-Robot Interaction (THRI)}, 8(3):1--24,
		2019.
		
		\bibitem{sachan2016easy}
		M.~Sachan and E.~P. Xing.
		\newblock Easy questions first? a case study on curriculum learning for
		question answering.
		\newblock In {\em Proceedings of the 54th Annual Meeting of the Association for
			Computational Linguistics}, volume~1, 2016.
		
		\bibitem{see2019what}
		A.~See, S.~Roller, D.~Kiela, and J.~Weston.
		\newblock What makes a good conversation? how controllable attributes affect
		human judgments.
		\newblock In {\em Proceedings of the 2019 Conference of the North}, 2019.
		
		\bibitem{cai2020learning}
		H.~Cai, H.~Chen, C.~Zhang, Y.~Song, X.~Zhao, Y.~Li, D.~Duan, and D.~Yin.
		\newblock Learning from easy to complex: Adaptive multi-curricula learning for
		neural dialogue generation.
		\newblock In {\em Proceedings of the AAAI Conference on Artificial
			Intelligence}, volume~34, pages 7472--7479, 2020.
		
		\bibitem{li2018microsoft}
		X.~Li, Y.~Wang, S.~Sun, S.~Panda, J.~Liu, and J.~Gao.
		\newblock Microsoft dialogue challenge: Building end-to-end task-completion
		dialogue systems.
		\newblock {\em arXiv preprint arXiv:1807.11125}, 2018.
		
		\bibitem{li2017user}
		X.~Li, Z.~C. Lipton, B.~Dhingra, L.~Li, J.~Gao, and Y.~Chen.
		\newblock A user simulator for task-completion dialogues.
		\newblock {\em arXiv preprint arXiv:1612.05688}, 2017.
		
	\end{thebibliography}
		
		
		
		
		% \end{thebibliography}
	
\end{document}